\documentclass{article}

\usepackage{PRIMEarxiv}

\usepackage[utf8]{inputenc} 
\usepackage[T1]{fontenc}    
\usepackage{hyperref}       
\usepackage{url}            
\usepackage{booktabs}       
\usepackage{amsfonts}       
\usepackage{nicefrac}       
\usepackage{microtype}      
\usepackage{lipsum}
\usepackage{fancyhdr}       
\usepackage{graphicx}       
\usepackage{comment}

\usepackage{lscape}
\usepackage{bm}
\usepackage{color}
\usepackage{ulem}
\usepackage{algorithm}
\usepackage{algorithmicx}
\usepackage{algpseudocode}
\usepackage{multicol}
\usepackage{amsmath,amssymb}

\newcommand{\NewDoubleCirc}{\hspace{.25em}\circ\hspace{-.8em}\bigcirc}
\newcommand{\ForAny}{\overset{\forall}{}}
\newcommand{\ForSome}{\overset{\exists}{}}

\newcommand{\activity}[1]{\textsf{#1}}
\newcommand{\anomaly}[1]{\textsf{#1}}
\newcommand{\ActSequences}[1]{\bm{#1}}

\title{Sensor Data Simulation for Anomaly Detection of the Elderly Living Alone}
\author{
  Kai Tanaka, Mineichi Kudo, Keigo Kimura \\
  Graduate School of Information Science and Technology \\
  Hokkaido University\\
  Sapporo, 060-0814, JAPAN\\
  \texttt{\{tanakai, mine, kimura5\}@ist.hokudai.ac.jp}
}

\begin{document}
\maketitle

\begin{abstract}
With the increase of the number of elderly people living alone around the world, there is a growing demand for sensor-based detection of anomalous behaviors. Although smart homes with ambient sensors could be useful for detecting such anomalies, there is a problem of lack of sufficient real data for developing detection algorithms. For coping with this problem, several sensor data simulators have been proposed, but they have not been able to model appropriately the long-term transitions and correlations between anomalies that exist in reality. In this paper, therefore, we propose a novel sensor data simulator that can model these factors in generation of sensor data. Anomalies considered in this study were classified into three types of \textit{state anomalies}, \textit{activity anomalies}, and \textit{moving anomalies}. The simulator produces 10 years data in 100 min. including six anomalies, two for each type. Numerical evaluations show that this simulator is superior to the past simulators in the sense that it simulates well day-to-day variations of real data.
\end{abstract}


\section{Introduction}
The rate of aging is globally increasing at a high rate. Indeed, the share of the elderly aged 65 years or older in the world was 9.3\% (727 million persons) in 2020, and is expected to reach over 16.0\% (1.5 billion persons) in 2050 \cite{UnitedNations2020WorldPopulationAgeing}. Moreover, in about 20 countries such as France and United Kingdom, the percentages of elderly persons who live alone exceed 30\% between 2006 and 2015 \cite{UnitedNations2020WorldPopulationAgeing}. For instance, Japan has the highest share of the older population (26.6\% in 2015) among major countries (whose total population is more than one million in 2015) \cite{He2016AnAgingWorld}, and about 18\% of the elderly people live alone in 2015 \cite{UnitedNations2020WorldPopulationAgeing}.

Aging is strongly connected to behavioral anomalies of elderly people. This is mainly because the functional and cognitive ability of a person start to decline by aging. For instance, 43.6\% of population aged 65 and older in 2019 in the United States have some disability, such as the mobility disability (27.7\% in the population) and the self-care disability (6.1\% in the population) \cite{CDC2019DisabilityandHealth}. As a measure of cognitive functioning, mini-mental state examination (MMSE) \cite{Folstein1975Mini-mentalstate} decreases by 4 points in median from the 18-24 age group to the 80+ age group \cite{Crum1993Population-basednormsfor}. These disabilities can cause dangerous or care-needed anomalies, including fall and wandering. In the worst case, the rate of patients who were found incapacitated or dead in their homes is increasing by age, e.g., 3 per 1000 among people aged 60 to 64, and 27 per 1000 among people aged 85 or older per year, as reported in San Francisco in 1993 \cite{Gurley1996Personsfoundin}.

The sustainable development goal (SDG) No. 3 also aims at ensuring healthy lives and promoting well-being for people at all ages. Also from this viewpoint, it is desirable to keep watching the health of elderly persons who live alone, and to maximize their healthy life expectancy. On the other hand, the world faces a lack of long-term care workers, especially in Japan \cite{Kondo2019Impactofincreased} and in many other developed countries \cite{Hussein2005Aninternationalreview}. Accordingly, it becomes crucial to answer the question who and how to care elderly people.

Smart home technology is possible to be an answer to the question \cite{Blackman2016Ambientassistedliving, Rashidi2013ASurveyon}. Smart homes are equipped with many ubiquitous sensors to monitor the bahavior of the resident. The sensor data can be used to monitor the health status and even to detect anomalies of the resident \cite{Cook2012HowSmartIs}. For instance, ubiquitous sensors including passive infrared motion sensors and door sensors are used to assess health status as mobility, cognition, or mood \cite{Alberdi2018SmartHome-BasedPrediction, Hayes2008Unobtrusiveassessmentof}. They are also used to detect deviating behaviors \cite{Lundstrom2016Detectingandexploring, Arifoglu2019Detectionofabnormal, Wong2014DataFusionwith} or concrete anomalies such as fall \cite{Tao2012Privacy-preservedbehavioranalysis} and wandering \cite{Zhao2014Alight-weightsystem, Gochoo2017Device-freenon-privacyinvasive, Chaudhary2020Sensorsignals-basedearly}.

The ultimate objective of this study is to develop algorithms to detect typical anomalies of elderly living alone at a smart home. However, in this paper, we focus on collecting sensor data. This is because, without a lot of sensor data, we cannot develop effective algorithms. More specifically, we need a lot of real sensor data including those of when anomalous behaviors happened. In addition, a large variety is desirable in the room layout, the age and the daily-based or monthly-based routine of living. However, in reality, it is almost impossible to collect real sensor data for four reasons; (1) a long-time, month to year, monitoring is necessary to collect sensor data, (2) the numbers of elderly people, the floor plans and the sensor arrangements are limited, (3) the occurrence of anomalies is uncontrollable, and (4) the labeling of behaviors and anomalies are very time-consuming and often difficult. As a result, developing sensor data simulators is necessary before developing anomaly detector algorithms. It is also desirable for the simulator to be independent from detection algorithms in a viewpoint of reality. This is the reason why we develop a sensor data simulator only in this paper.

Recently, many sensor data simulators have been proposed \cite{Synnott2015SimulationofSmart}. Some of them produce even anomalous patterns \cite{Arifoglu2019Detectionofabnormal, Moallem2019AnomalyDetectionin, Lundstrom2016Detectingandexploring, Alshammari2018SIMADL:SimulatedActivities, Bouchard2010SIMACT:A3D, Francillette2020Modelingthebehavior, Masciadri2018DisseminatingSyntheticSmart, Jiang2021SISG4HEI_Alpha:Alphaversion}. However, they are not sufficient in the sense that they do not consider the facts that (1) anomalies get increase and/or worse over time, and that (2) there exists the correlation between anomalies. As for (1), we need to incorporate such a mechanism that increase the frequency of some anomalies as time goes. As for (2), for example, it is known that dementia can cause some anomalies \cite{Muller2022Behavioraldisturbancesin}. To cope with these insufficiency, we propose a novel activity/sensor data simulator. The contribution of this paper is as follows.

\begin{description}
\item[(1)] The proposed simulator covers six typical anomalies that are followed by the physical movement of the resident and, thus, can be detected by several binary sensors such as infrared motion or pressure sensors. They are two from each of three categories: \textit{state anomalies} (\anomaly{semi-bedridden}, \anomaly{housebound}) over many days, \textit{activity anomalies} (\anomaly{wandering}, \anomaly{forgetting to turn off the home appliances}) occasionally happening, \textit{moving anomalies} (\anomaly{fall while standing}, \anomaly{fall while standing}) happening during movements.

\item[(2)] The simulated anomalies range from seconds to years, unlike previous simulators that deal with only either short-term or long-term anomalies. To realize this long range, we consider a graphical model that represents a hierarchical causal relationship among anomalies, especially from long-term activities to short-term activities such as walking steps.

\item[(3)] The correlation among anomalies is statistically modeled with a latent variable of MMSE. Therefore, the frequency of anomalies or the period of the status becomes higher or longer according to the progress of symptoms of dementia with age.

\item[(4)] In activity sequence comparison, it is shown that the activity sequences generated by the proposed simulator are sufficiently close to real ones.
\end{description}

A preliminary version of this study has been reported in \cite{Tanaka2022SensorDataSimulation}. This paper is extended from \cite{Tanaka2022SensorDataSimulation} in detailed survey (Tables \ref{tab:comparison_between_sensor_data_simulations} and \ref{tab:comparison_of_simulated_anomaly}), the statistic model (Tables \ref{tab:parameters_of_each_activity} and Fig. \ref{fig:graphical_model}), the number of anomalies (from one to six), and qualitative and quantitative evaluations (Sections \ref{sec:qualitative_evaluation} and \ref{sec:quantitative_evaluation}).

\section{Related Works}

\begin{table}
\centering
\caption{Comparison among sensor data simulators with anomalous patterns. Abbreviations are; simulation speed (SS), environment information such as the room layout and the sensor arrangement (EI), human activity on how a resident behaves a day and informations about walking trajectories (HA), anomalous behavior (AB), and information to reproduce same statistical properties for data (IRSSPD). Evaluations are; $\NewDoubleCirc$: Exact, $\bigcirc$: Approximate, and $\times$: Not available.}
\label{tab:comparison_between_sensor_data_simulations}
\scalebox{1}[1]{
\begin{tabular}[tbp]{p{3.8cm}p{3.3cm}p{0.5cm}p{0.1cm}p{0.4cm}p{0.4cm}p{0.4cm}p{0.1cm}p{3.5cm}}
\\
\hline
Type & Method [Ref.] & SS & & \multicolumn{3}{c}{Adjustability} & & IRSSPD\\ \cline{5-7}
& & & & EI & HA & AB & &\\
\hline
Manual modification & \cite{Arifoglu2019Detectionofabnormal}, \cite{Moallem2019AnomalyDetectionin}, \cite{Lundstrom2016Detectingandexploring} & low & & $\times$ & $\times$ & $\NewDoubleCirc$ & 
& real data and inserted anomaly patterns\\\\
Manipulating an avatar & \cite{Alshammari2018SIMADL:SimulatedActivities} & low & & $\bigcirc$ & $\NewDoubleCirc$ & $\NewDoubleCirc$ & & manipulator's habit\\\\
Agent scenario generation & \cite{Bouchard2010SIMACT:A3D}, \cite{Francillette2020Modelingthebehavior}, \cite{Masciadri2018DisseminatingSyntheticSmart}, \cite{Kristiansen2016Smoothandcrispy}, \cite{Jiang2021SISG4HEI_Alpha:Alphaversion}, Proposed & high & & $\bigcirc$ & $\bigcirc$ & $\bigcirc$ & & scripts of models or hyperparameters of models\\
\hline
\end{tabular}
}
\end{table}

Simulated sensor data are usually collected from virtual sensors attached to everywhere in the virtual smart home, according to the behavior of a virtual agent. We start with a survey on the historical stream of such sensor data simulators, especially paying attention to how anomaly patterns are incorporated in such simulators.

Sensor data simulators have been realized in several ways \cite{Synnott2015SimulationofSmart, Golestan2022SmartIndoorSpace}, some of which consider anomalies \cite{Arifoglu2019Detectionofabnormal, Moallem2019AnomalyDetectionin, Lundstrom2016Detectingandexploring, Alshammari2018SIMADL:SimulatedActivities, Bouchard2010SIMACT:A3D, Francillette2020Modelingthebehavior, Masciadri2018DisseminatingSyntheticSmart, Kristiansen2016Smoothandcrispy, Jiang2021SISG4HEI_Alpha:Alphaversion}. Anomaly implementations are classified into three types: \textit{Manual modification} \cite{Arifoglu2019Detectionofabnormal, Moallem2019AnomalyDetectionin, Lundstrom2016Detectingandexploring}, \textit{Manipulating an avatar} \cite{Alshammari2018SIMADL:SimulatedActivities} and \textit{Agent scenario generation} \cite{Bouchard2010SIMACT:A3D, Francillette2020Modelingthebehavior, Masciadri2018DisseminatingSyntheticSmart, Jiang2021SISG4HEI_Alpha:Alphaversion}. We summarize them in Tables \ref{tab:comparison_between_sensor_data_simulations} and \ref{tab:comparison_of_simulated_anomaly}. Table \ref{tab:comparison_between_sensor_data_simulations} compares important factors in each simulator type. Table \ref{tab:comparison_of_simulated_anomaly} shows what kind of anomalous behaviors are considered and how they are generated. Note that none of them consider both a gradually changing frequency/duration of anomalies and the correlation among anomalies.

Let us describe the methods in detail type by type.

\subsubsection{Manual modification}\label{sec:manual_modification}
In the type of manual modification, anomaly patterns are inserted artificially into real sensor data collected from a $real$ smart home. For instance, Arifoglu and Bouchachia \cite{Arifoglu2019Detectionofabnormal} inserted sensor data corresponding to repetitive behaviors and sleep disruption, by considering anomalous behavioral symptoms of a person with dementia. Moallem, Hassanpour and Pouyan \cite{Moallem2019AnomalyDetectionin} also inserted sensor data corresponding to the events that rarely happen in real data. They also added sensor data of a temporal anomaly of which duration time varies. Lundstr\"{o}m, J\"{a}rpe and Verikas \cite{Lundstrom2016Detectingandexploring} considered some combinations of temporal, transitional and spatial deviations of anomalous behaviors, for example, the resident moves after a fall at bathroom and watch TV at night. The problem of this type is the necessity of long-time monitoring. It also has less variety of room-layout, sensor arrangement and anomalies.

\subsubsection{Manipulating an avatar}
This type of simulators uses an avatar, a virtual model of a real person, instead of a real elderly person. The behavior of the avatar is manually controlled by some subjects. The avatar interacts with the objects such as a door and a TV in the house. Alshammari \textit{et al.} \cite{Alshammari2018SIMADL:SimulatedActivities} let seven subjects to use the simulator \cite{Alshammari2017OpenSHS:Opensmart} to reproduce behavioral anomalies, e.g., leaving the fridge door open for a long time. The merit is that there is no risk to injure the real resident due to anomalies. This type has the same long-time monitoring problem as the manual modification type. 

\subsubsection{Agent scenario generation}
This type of simulators prepares some intermediate data for controlling how an agent acts autonomously in a virtual smart home. The intermediate data specify the daily schedule of the agent, a walking model of the agent, and the activation rule of sensors. This type generates much sensor data faster than the previous two types, but the control is more difficult. Bouchard \textit{et al.} \cite{Bouchard2010SIMACT:A3D} control the resident's activities by scripts. They wrote a scenario even for an agent with dementia, but the detail is unfortunately not shown. Because of manually written scripts, the anomalous patterns are limited in number and in variety. Moreover, the behavior of the agent is specified by the script writer. Francillette \textit{et al.} \cite{Francillette2020Modelingthebehavior} randomly and automatically inserted a wider range of anomalous patterns into sensor data generated by a sensor data simulator \cite{Francillette2017TheVirtualEnvironment}: omission of activities, substitution of activities, wrong order between activities, perseveration of activities, and addition of irrelevant sub-actions in activities. They are symptoms of a person with mild cognitive impairment of the mild Alzheimer's disease. However, they did not deal with long-term anomalies such as being housebound. In addition, it is sometimes observed that generated frequency of anomalies are unrealistically high or low. Veronese \textit{et al.} \cite{Masciadri2018DisseminatingSyntheticSmart} generated and opened in public sensor data that are based on the sensor data generator SHARON \cite{Veronese2017Realistichumanbehaviour} and took into consideration the patients of dementia. It succeeded in modeling transitional patterns such as interrupted sleep, increased duration time of complex behaviors, and a reduced frequency of dinner. Casaccia \textit{et al.} \cite{Casaccia2020MeasurementofActivities} firstly simulated wandering, that is an anomalous travel pattern while walking in a room. However, they only paid attention to the trajectory patterns of wandering, not its frequency or duration over days. Jiang and Mita \cite{Jiang2021SISG4HEI_Alpha:Alphaversion} also considered wandering. However, the frequency was fixed and the duration time between two wanderings were also almost fixed, due to the activity schedule generator \cite{Jiang2020AutomaticDailyActivity}. This way cannot simulate an increase of times of wandering, as age goes, and an unpredictable occurrence of wandering. Kristiansen, Plagemann and Goebel \cite{Kristiansen2016Smoothandcrispy} simulated sensor data with an activity sequence generator \cite{Kristiansen2018AnActivityRule}. However, their virtual sensor outputs are obtained only from an activity sequence without the behavior of an agent, independent of the room layout \cite{Kristiansen2016Smoothandcrispy}. They considered also hypothetical anomalous behaviors, such as a gradual onset of sleep disorder (e.g., shifting start time of sleep, and increasing duration time of sleep) and a gradual improvement of self-medication routine to take medicines at the accurate time. Their simulation loses a reality in the sense that the virtual sensors directly capture the designed behavior, without observation of the agent's physical movements.

In summary, previous simulators have at least one of the following problems:
\begin{description}
\setlength{\itemsep}{1pt}
\setlength{\leftskip}{0.5cm}
\item[S1] A long-term trend of statistical properties of anomalies, typically moving worse as the age goes, cannot be handled,
\item[S2] Correlations among anomalies are not considered, in spite of the fact that several anomalies tend to happen more frequently together as the age goes.
\item[S3] Sensor data are not always collected from sensors response activated by the physical movement (walking near it or using it) of the agent.
\end{description}

To cope with S1 and S2, we consider a latent variable representing the degree of dementia varying over time, because dementia often gets worse with age \cite{Crum1993Population-basednormsfor}. Also to cope with S2, we adopt a statistical graphical model where each anomaly is governed by the latent variable in frequency and/or in duration time. To cope with S3, we use the sensor response that are activated by the resident's movements in the house.

\begin{landscape}
\begin{table}
\centering
\caption{Summary of simulators to generate anomalous patterns. Bold faces indicate that the anomaly is simulated only in the reference.}
\label{tab:comparison_of_simulated_anomaly}
\begin{tabular}[tbp]{p{1.6cm}p{0.8cm}p{2.6cm}p{2.5cm}p{3.2cm}p{6cm}p{0.5cm}p{0.4cm}}
\\
\hline
Type & Ref. & Floor plan * & Activity * & Sensor *, ** & Simulated anomalies (the number)& \multicolumn{2}{c}{Viewpoint ***}\\
\cline{7-8}
 & & &  & & & LT & CR\\
\hline
Manual modification & \cite{Arifoglu2019Detectionofabnormal} & RD (\cite{Cook2013CASAS:ASmart, Cook2009Assessingthequality}) & mRD & mRD & \textbf{repetitive behaviors}, sleep disruption, confusion (3) & $\times$ & $\times$\\
           & \cite{Moallem2019AnomalyDetectionin} & RD (\cite{Cook2013CASAS:ASmart}) & mRD & mRD & frequency anomaly, temporal anomaly, transitional anomaly (3) & $\times$ & $\times$\\
                    & \cite{Lundstrom2016Detectingandexploring} & RD (\cite{Eriksson2011Tryggomnatten:}) & mRD & mRD & spatial anomaly, temporal anomaly, transitional anomaly (3) & $\times$ & $\times$\\
\hline
Manipulating an avatar & \cite{Alshammari2018SIMADL:SimulatedActivities} & 3D layout made with GUI & determined by a user & DOOR, PIR, SWITCH & anomalous control patterns of home appliances (1) & $\times$ & $\times$\\
\hline
Agent scenario generation & \cite{Bouchard2010SIMACT:A3D} & 3D layout & scripts & virtual sensor represented by scripts of objects & (unclear) & $\times$ & $\times$\\
                          & \cite{Francillette2020Modelingthebehavior} & 3D layout made with GUI & generated by behavior trees & COST, PIR, PR, RFID and contact sensor & \textbf{omission of activities}, \textbf{substitution of activities}, \textbf{wrong order between activities}, \textbf{perseveration of activities}, \textbf{addition of irrelevant sub-actions in activities} (5) & $\times$ & $\times$\\
                          & \cite{Masciadri2018DisseminatingSyntheticSmart} & 2D layout made by a text & motivation based planning with habitual activities & DOOR, COST, PIR, SWITCH and motion sensor & interrupted sleep, \textbf{increased duration time of complex behaviors}, \textbf{a reduced frequency of dinner and going out} (3) & $\bigcirc$ & $\times$\\
                          & \cite{Kristiansen2016Smoothandcrispy} & (unclear) & determined by a distribution with priority rules & time, position and logical sensors & immobility caused by falling, sleep disorder (2) & $\bigcirc$ & $\times$\\
                          & \cite{Jiang2021SISG4HEI_Alpha:Alphaversion} & automatic generation of 2D layout and height data of studio apartment & probability motivation based planning & PIR & wandering (1) & $\times$ & $\times$\\\\
                          & {\small This paper} & borrowed from \cite{Jiang2021SISG4HEI_Alpha:Alphaversion} with some modifications & determined by a distribution with hierarchical order & COST, PIR, PR & \anomaly{falling while walking}, \textbf{\anomaly{falling while standing}}, \anomaly{wandering}, \anomaly{forgetting}, \textbf{\anomaly{housebound}}, \textbf{\anomaly{semi-bedridden}} (6) & $\bigcirc$ & $\bigcirc$\\
\hline
\end{tabular}
\\
\flushleft{* RD: real data, mRD: modified real data, 2(3) D layout: 2(3) dimensional artificially designed layout.\\
** COST: cost (power or flow) sensor, DOOR: door sensor, PIR: passive infrared motion sensor, PR: pressure sensor, RFID: radio-frequency identification based sensor, and SWITCH: switch of home appliances.\\
*** LT: whether the method controls long-term trends of statistical properties of anomalous behaviors, CR: whether or not the method considers the causal relationships among multiple anomalous behaviors ($\bigcirc$: equipped, and $\times$: not equipped).}
\end{table}
\end{landscape}

\section{Sensor data simulation}\label{sec:sensor_data_simulation}

The terminology of this paper is shown in Table \ref{tab:symbols_to_use_definitions}.

\begin{table}
\centering
\caption{Symbols to use definitions.}
\label{tab:symbols_to_use_definitions}
\begin{tabular}[tbp]{p{3.5cm}p{6.5cm}p{5cm}}
\hline
Symbol & Content & Example\\ \hline
$I_b=\{1, 2, \ldots, N_b\}$ & Index set of binary sensors. & -\\
$I_s=\{1, 2, \ldots, N_s\}$ & Index set of sampling. & -\\
$T_s$ & Sampling period (constant in common to all sensors). & -\\
$f_t\colon I_s\rightarrow\mathbb{R}_{+}$ & Elapsed time of $i$th sampling $f_t(i) = T_s\cdot i$. & $f_t(3) = 10\textrm{d}~08\textrm{h}~55\textrm{m}~30.0\textrm{s}$\\
$S_A=S_F\cup S_N\cup S_R$ & Set of (Fundamental, Necessary, and Random) activities. & $\{\activity{sleep}, \activity{urination}, \ldots, \activity{have dinner}\}$\\
$U\subseteq\mathbb{R}^{2}$ & Possible grid locations in the house. & $\{(1.0, 1.0), \ldots, (1000.0, 500.0)\}$\\
$f_z\colon S_A\rightarrow 2^U$ & Allocated zone for the activity. & $f_z(\activity{sleep}) = \{(5, 5), \ldots, (60, 30)\}$\\
$A = (\bm{a}_1, \ldots, \bm{a}_{N_A})$ & Activity sequence over days. & -\\
$\bm{a}_k= (a_{k,n}, a_{k,s}, a_{k,d}, a_{k, l}) $ & The $k$th activity (activity $a_{k,n}$ starts at $a_{k, s}$ for duration $a_{k,d}$ at location $a_{k, l}$). & 
$\left(\activity{sleep}, 3\textrm{d}~22\textrm{h}~13\textrm{m}, 430, (50, 30)\right)$\\

$W_{\bm{a}, \bm{b}}\in\left(\mathbb{R}\times U\right)^{+}$ & Walking trajectory when activity changes from $\bm{a}$ to $\bm{b}$ ($+$ is the Kleene closure). & \mbox{$W_{\bm{a}, \bm{b}}=(~(9\textrm{h}~6\textrm{m}~3\textrm{s}, (500, 40))$}, $\ldots, (9\textrm{h}~7\textrm{m}~16\textrm{s}, (700, 40))~)$\\

$M_m$ & Random variable of the MMSE score in $m$th [month] with a decreasing trend. & $M_1$ = 29\\
$F_{a, m}~(D_{a, m})$ & Mean frequency (duration) of anomaly $a$ in $m$th [month]. & $F_{\anomaly{forgetting}, m} = 3 \textrm{[times/month]}$,\par\noindent $D_{\anomaly{wandering}, m} = 2 \textrm{[min.]}$\\
\hline
\end{tabular}
\end{table}

Let $N_b\in\mathbb{N}$ be the number of binary-output sensors such as passive infrared motion sensors and pressure sensors, and $I_b=\{1, 2, \ldots, N_b\}$ be the index set of them. All sensors are sampled with period $T_s$. For now, $T_s$ is assumed to be in common to all sensors, for simplicity. Let $N_s\in\mathbb{N}$ be the number of times of sampling, and $I_s=\{1,2,\ldots, N_s\}$ be the index set of sampling. The $i$th sampling time $f_t(i)$ is calculated by $f_t(i) = T_s\cdot i$. We define sensor data as follows. Let $d_{i, j}\in\{0, 1\}$ be the state of binary sensor $j\in I_b$ at sampling point $i\in I_s$. We define a sensor state vector at $i\in I_s$ as $\bm{d}_i = [d_{i, 1}, d_{i, 2}, \ldots, d_{i, N_b}]^{T}$, where $T$ means the transpose. In total of $N_s$ samplings, we have a data matrix $D = [\bm{d}_1, \bm{d}_2, \ldots, \bm{d}_{N_s}]$. For reduction of data volume, we just keep successive different state vectors only, that is $R = [\bm{d}_{i_1}, \bm{d}_{i_2}, \ldots, \bm{d}_{i_N}]$ where $\bm{d}_{i_k}\neq \bm{d}_{i_{k+1}}$.

\subsection{Four component simulators}
\begin{figure}[tbp]
\centering
\includegraphics[width=150mm]{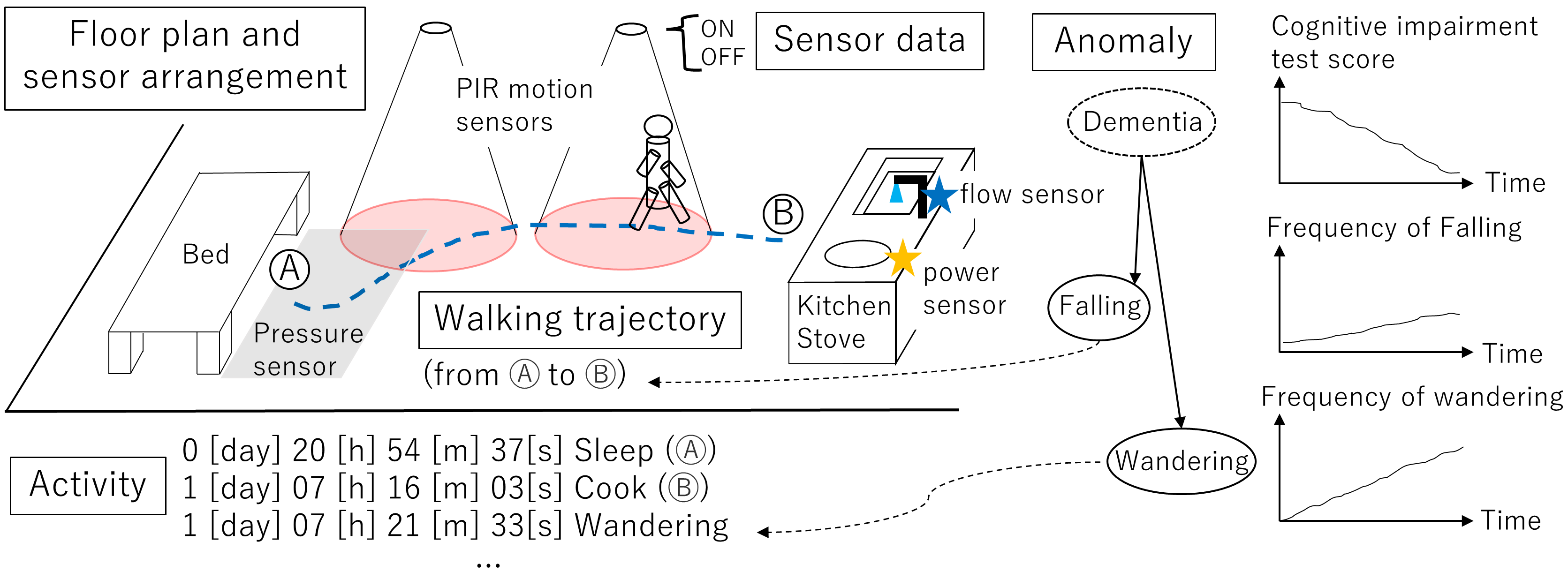}
\caption{The proposed simulator consisting of four component simulators and a sensor data collector (each surrounded by lines).}
\label{fig:abstract_of_simulator}
\end{figure}

The proposed simulator is an extension of that in \cite{Jiang2021SISG4HEI_Alpha:Alphaversion}, an agent scenario generation method. It is divided into mainly three component simulators \cite{Jiang2021SISG4HEI_Alpha:Alphaversion, Lee2016Automaticagentgeneration, Francillette2017TheVirtualEnvironment}: (1) a floor plan simulator, (2) an activity simulator, and (3) a walking trajectory simulator. We add one more component, (4) an anomalous behavior simulator. The relationship among those component simulators is illustrated in Fig. \ref{fig:abstract_of_simulator}. Each component simulator is explained as follows. Here, the arrangement of sensors is included in (1).

\subsubsection{Floor plan simulator \cite{Jiang2021SISG4HEI_Alpha:Alphaversion}}
A floor plan simulator determines the layout of rooms and furniture. It also determines the positions and kinds of sensors. We borrowed the floor plan simulator of studio apartments from \cite{Jiang2021SISG4HEI_Alpha:Alphaversion}. In the simulator, rooms and furniture are automatically set according to rules that are recommended for elderly persons, for example, a piece of furniture must be apart 0.6~[m] from the others \cite{Jiang2021SISG4HEI_Alpha:Alphaversion}. Positions of sensors are determined manually or randomly by specifying their two dimensional coordinates.

\begin{figure}[tbp]
\centering
\includegraphics[width=120mm]{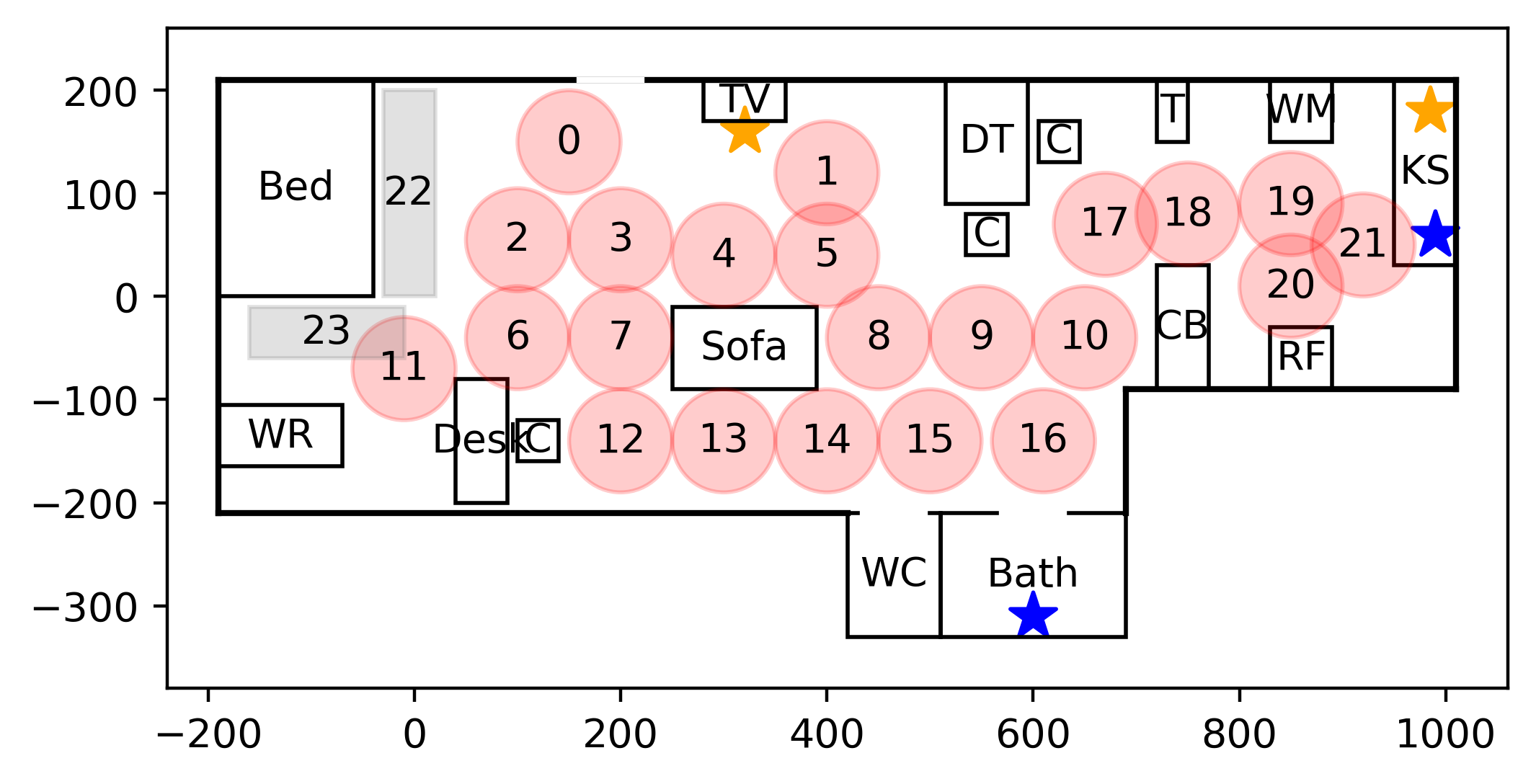}
\caption{Floor plan and sensor arrangement. The unit of length is a meter. There are chair (C), cupboard (CB), dining table (DT), kitchen stove (KS), refrigerator (RF), trash box (T), wardrobe (WR), washing machine (WM) and water closet (WC). Colors and symbols distinguish kinds of sensors; blue for flow COST sensors, yellow for power COST sensors, gray for PR sensors, and red for PIR sensors.}
\label{fig:floor_plan_and_sensor_arrangement}
\end{figure}

In this study, in the viewpoint of privacy protection and avoiding annoyance of sensor attachment to a body, we consider mainly three kinds of binary sensors of (1) passive infrared (PIR) motion sensors, (2) pressure (PR) sensors, and (3) cost (COST) sensors (power sensors and flow sensors). In the experiment, PIR motion sensors were placed manually so as to cover the main walking flow of the resident. PR sensors were placed at the foot side of the bed. Among COST sensors, power sensors were attached to a TV set and a kitchen stove, and liquid flow sensors were attached at kitchen and bathroom (Fig. \ref{fig:floor_plan_and_sensor_arrangement}).

\subsubsection{Activity simulator}
An activity simulator generates a time series of activities that represent the daily behavior of the resident. 

We define an activity sequence $A=(\bm{a}_1, \bm{a}_2, \ldots, \bm{a}_{N_A})$ across multiple days as follows, where $\bm{a}_k$ is the $k$th activity. The $k$th activity $\bm{a}_k = [a_{k,n}, a_{k,s}, a_{k, d}, a_{k, l}]^{T}$ is composed of the name $a_{k, n}\in S_{A}$, the starting time $a_{k, s}\in\mathbb{R}$, the duration time $a_{k, d}\in\mathbb{R}$, and the starting location $a_{k, l}\in U\subseteq\mathbb{R}^2$, where $U$ is the set of possible locations on the grid base design in the house. The activity $\bm{a}_k$ starts at $a_{k, s}$ and continues until $a_{k, s} + a_{k, d}$, for example, $\bm{a}_i = $(\activity{have lunch}, 2d 12h 10m, 56m, (10, 20)) means that the resident eats lunch from 12:10 p.m. to 13:06 p.m. on day two at position (10, 20). We assume that each activity is acted at a certain place connected to the activity in the house, for example, \activity{read a book} is made at desk or at sofa. So we define $f_z\colon S_A\rightarrow 2^U$ as the function that determines possible start locations of an activity. The concrete location of $a_{k, l}$ is chosen at random from $f_z(a_{k, n})$. The sequence $A$ is required to cover a day without gap.

\begin{figure}[tbp]
\centering
\includegraphics[width=140mm]{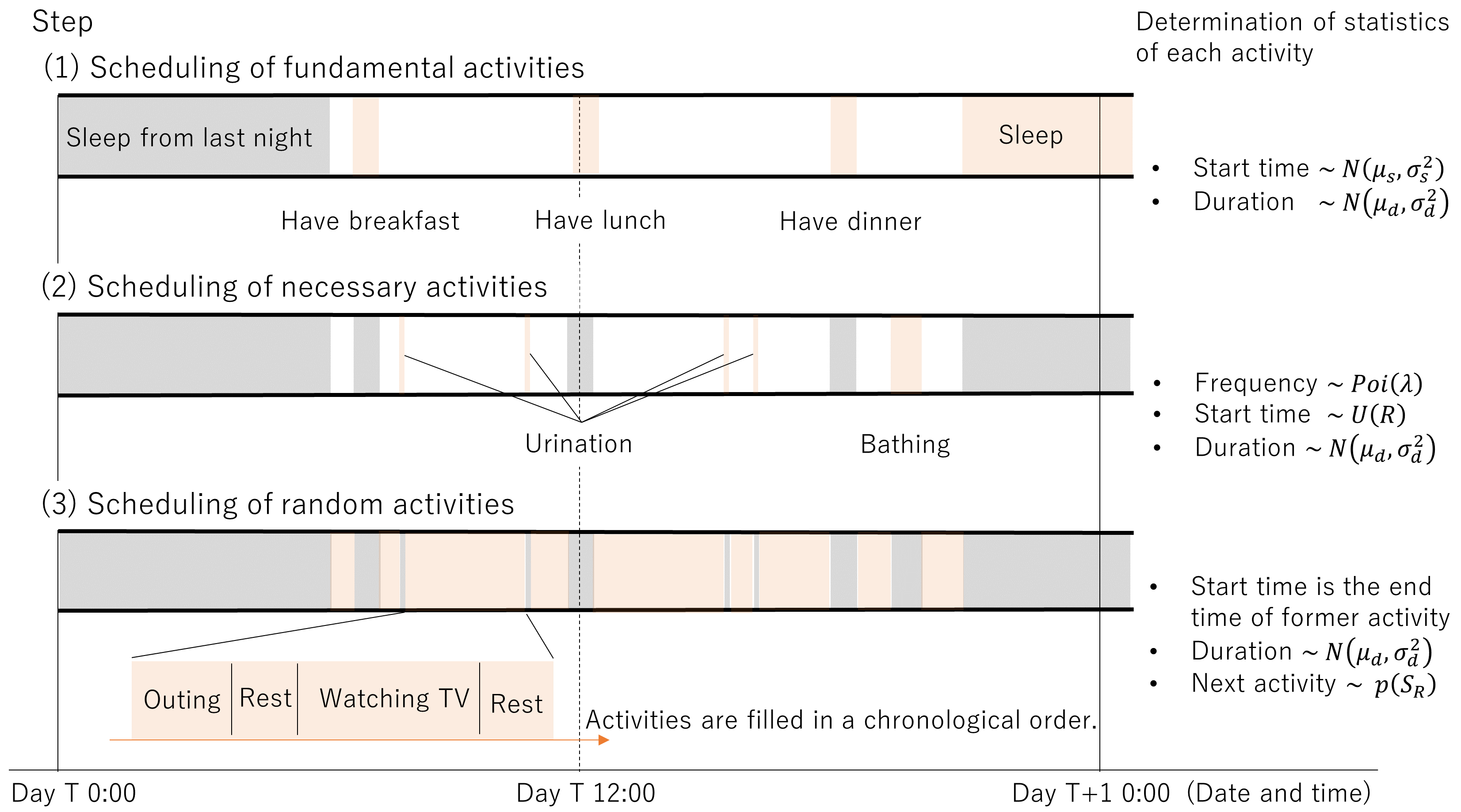}
\caption{Activity scheduling of a day. (1) Fundamental, (2) necessary, and (3) random activities are detemrined in the order. Their start times, durations and other statistics are determined by probabilistic models: a normal distribution $N(\mu, \sigma^2)$ with mean $\mu$ and variance $\sigma^2$, a Poisson distribution $Poi(\lambda)$ with mean $\lambda$, an uniform distribution $U(R)$ with remainder time range $R$ that is not filled yet, and a distribution $p(S_R)$ of occurrence probabilities over random activities $S_R$.}
\label{fig:activity_scheduling}
\end{figure}

We developed a novel activity simulator of normal activities. It divides daily activities into three types of $S_F$: fundamental activities, $S_N$: necessary activities, and $S_R$: random activities, so that $S_A = S_F\cup S_N\cup S_R$. The proposed algorithm fills out a day by activities of those types in order of $S_F$, $S_N$ and $S_R$. We introduce a flexible probability distribution in random activities. One instance is shown in Fig. \ref{fig:activity_scheduling}. The algorithm is shown in \ref{sec:algorithm_of_activity_scheduling}. The filling procedure is as follows. 
\begin{description}
\setlength{\itemsep}{1pt}
\setlength{\leftskip}{0.5cm}
\item[Fundamental activities]  of $S_F$ (e.g., \activity{have lunch}, \activity{sleep}) are activities that happen necessarily at least once a day for keeping independent life. Each is specified by the starting time $a_s$ and the duration time $a_d$ that are statistically modeled by $a_s\sim \mathcal{N}(\mu_{a, s}, \sigma_{a, s}^2)$ and $a_d\sim \mathcal{N}(\mu_{a, d}, \sigma_{a, d}^2)$, where $\mathcal{N}$ is a normal distribution. 
\item[Necessary activities] of $S_N$ (e.g., \activity{change clothes}, \activity{urination}) are activities that are almost necessary for keeping independent life. Each can happen multiple times a day and is specified by the occurrence frequency $\lambda_{a}$ of a day and the duration time $a_d$ where $\lambda_{a}\sim \mathrm{Poi}(\mu_{a, d})$ and $a_d\sim \mathcal{N}(\mu_{a, d}, \sigma_{a, d}^2)$, where $\mathrm{Poi}$ is a Poisson distribution. 
\item[Random activities] of $S_R$ (e.g., \activity{go out}, \activity{watch TV}) are activities that are not always necessary but desirable for keeping independent life. It is specified by the duration time $a_d$ and the occurrence probability $p_a$ in $S_R$ where $a_d\sim \mathcal{N}(\mu_{a, d}, \sigma_{a, d}^2)$ and $\sum_{a\in S_R}{p_a} = 1$. For every $a\in S_R$, the values of $p_a$ was manually set up or learned by data.
\end{description}
In summary, $a\in S_F$ is specified by $(\mu_{a, s}, \sigma_{a, s}^2, \mu_{a, d}, \sigma_{a, d}^2)$, $a\in S_N$ is specified by $(\lambda_{a}, \mu_{a, d}, \sigma_{a, d}^2)$, and $a\in S_R$ is specified by $(\mu_{a, d}, \sigma_{a, d}^2, p_a)$. The parameters used in the experiments are shown in Table \ref{tab:parameters_of_each_activity}.

\begin{table}
\centering
\caption{Classification of 20 activities (F: Fundamental, N: Necessary, R: Random) and their statistics. Here, $\mu_{a, s}$ and $\sigma_{a, s}$ denote the mean and the standard deviation of start time, $\mu_{a, d}$ and $\sigma_{a, d}$ denote those of duration time, and $\lambda_{a}$ is the mean frequency in a day. The unit is a minute. The values are determined from statistical surveys in Japan \cite{StatisticsBureauofJapan2016StatisticsBureauHome}.}
\label{tab:parameters_of_each_activity}
\begin{tabular}[tbp]{p{0.5cm}p{2.6cm}p{1.9cm}p{0.5cm}p{2.5cm}p{0.5cm}p{0.5cm}p{1.5cm}p{1.5cm}}
\\
\hline
Type & Activity & \multicolumn{2}{c}{Start time [m]} & Freq. [times/day] & \multicolumn{2}{c}{Duration [m]} & Place & Appliances\\
\cline{3-4}\cline{6-7}
& & $\mu_{a, s}$ & $\sigma_{a, s}$ & $\lambda_{a}$ & $\mu_{a, d}$ & $\sigma_{a, d}$ & & \\ \hline
\textbf{F} & \activity{have breakfast} & 437 (7:17) & 30 & --- & 34 & 10 & Table & ---\\
           & \activity{have lunch} & 720 (12:00) & 30 & --- & 42 & 10 & Table & ---\\
           & \activity{have dinner} & 1098 (18:18) & 30 & --- & 45 & 10 & Table & ---\\
           & \activity{sleep} & 1289 (21:29) & 40 & --- & 482 & 30 & Bed & ---\\ \hline
\textbf{N} & \activity{brush teeth} & --- & --- & 2 & 1.5 & 0.5 & Bathroom & Faucet\\
           & \activity{change clothes} & --- & --- & 2 & 5 & 1 & Wardrobe & ---\\
           & \activity{defecation} & --- & --- & 1 & 10 & 3 & Toilet & ---\\
           & \activity{take a bath} & --- & --- & 1 & 30 & 10 & Bathroom & Shower\\
           & \activity{urination} & --- & --- & 5 & 3 & 0.5 & Toilet & ---\\
           & ~~~~$\anomaly{wandering}^{*w}$ & --- & --- & $\lambda_w$ & $\mu_w$ & $\mu_w/5$ & Walking & ---\\
           & ~~~~$\activity{go out}^{*h}$ & --- & --- & $1/14$ & 20 & 4 & Entrance & ---\\
           & ~~~~~$\activity{use the phone}^{*h}$ & --- & --- & $1/3$ & 10 & 2 & Desk & ---\\
           & ~~~~$\activity{go out}^{*s}$ & --- & --- & $1/7$ & 20 & 4 & Entrance & ---\\
\textbf{R} & \activity{clean} & --- & --- & --- & 30 & 10 & Trash box & ---\\
           & \activity{cooking} & --- & --- & --- & 30 & 10 & Kitchen & Stove\\
           & $\activity{go out}$ & --- & --- & --- & 40 & 20 & Entrance & ---\\
           & \activity{read a book} & --- & --- & --- & 40 & 20 & Desk & ---\\
           & $\activity{rest}$ & --- & --- & --- & 30 & 10 & Sofa & ---\\
           & \activity{take a snack} & --- & --- & --- & 10 & 3 & Table & ---\\
           & $\activity{use the phone}$ & --- & --- & --- & 10 & 3 & Desk & ---\\
           & \activity{wash clothes} & --- & --- & --- & 5 & 1 & Washer & ---\\
           & \activity{watch TV} & --- & --- & --- & 40 & 20 & Sofa & TV\\
           & ~~~~$\activity{nap}^{*s}$ & --- & --- & --- & 40 & 8 & Sofa & ---\\
           & ~~~~$\activity{rest}^{*s}$ & --- & --- & --- & 60 & 10 & Sofa & ---\\
\hline
\end{tabular}
\flushleft{
*w: the frequency $\lambda_w$ and duration $\mu_w$ of \anomaly{wandering} are determined depending on anomaly simulators.\\
*h: during \anomaly{housebound}, \activity{go out} and \activity{use the phone} are switched from random to necessary type.\\
*s: during \anomaly{semi-bedridden}, \activity{go out} is switched from random to necessary type, the duration of \activity{rest} increases, and \activity{nap} appears in random activities.\\
**hs: when \anomaly{housebound} and \anomaly{semi-bedridden} happen at the same time, \activity{go out} with *s is applied.\\
}
\end{table}

\subsubsection{Walking trajectory simulator \cite{Jiang2021SISG4HEI_Alpha:Alphaversion, Jiang2020Automaticspatialattribute}}
\begin{figure}[tbp]
\centering
\includegraphics[width=\linewidth]{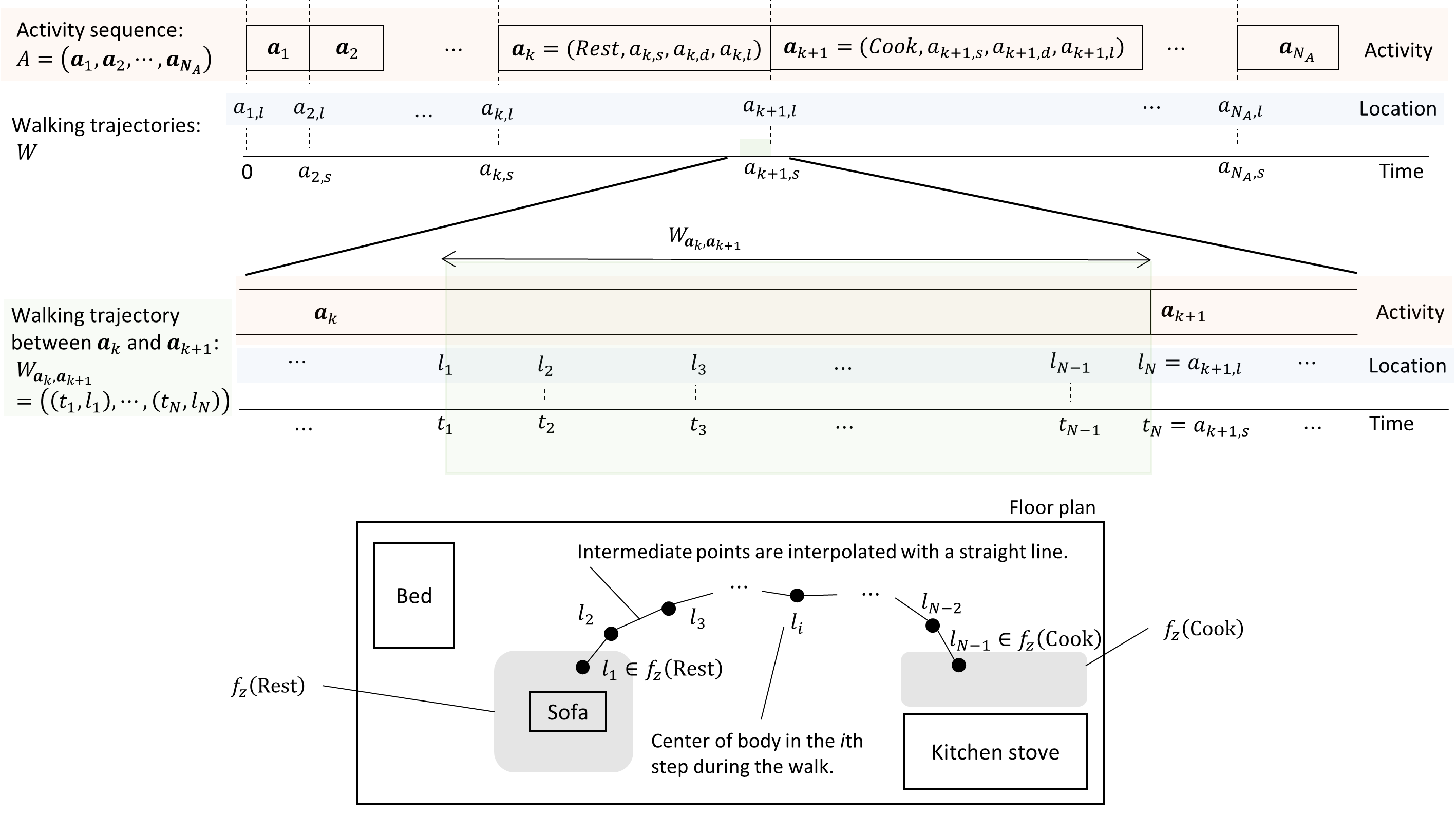}
\caption{Generation of a walking trajectory from an activity sequence. A walking trajectory between $\bm{a}_k$ and $\bm{a}_{k+1}$ is $W_{\bm{a}_k, \bm{a}_{k+1}} = \left((t_1, l_1), (t_2, l_2), \ldots, (t_N, l_N)\right)$, where $l_1 = a_{k, l}$ and $l_N = a_{k+1, l}$. The trajectory is obtained by connecting body centers in the movement almost straight while avoiding furniture. Body centers are determined in short and smooth connected steps with a specified stride length.}
\label{fig:illustrated_symbols}
\end{figure}

In the agent scenario method, an agent moves from the current position to the next position autonomously to do the next activity. We borrowed a walking trajectory simulator proposed in \cite{Jiang2021SISG4HEI_Alpha:Alphaversion, Jiang2020Automaticspatialattribute}. A walking trajectory simulator generates a moving path (Fig. \ref{fig:illustrated_symbols}).

The trajectories are determined as follows. Let $W_{\bm{a}, \bm{b}}\in(\mathbb{R}\times U)^{+}$ be a sequence of time and location pairs between the current activity $\bm{a}$ and the next activity $\bm{b}$, where $+$ is the Kleene closure. Here, each location shows the center of the resident's body. For example, when he/she changes the activity from $\bm{a}_k = (a_{k, n}, a_{k, s}, a_{k, d}, a_{k, l})$ to $\bm{a}_{k+1} = (a_{k+1, n}, a_{k+1, s}, a_{k+1, d}, a_{k+1, l})$, the trajectory, with duration time $T_W$, is given by $W_{\bm{a}_k, \bm{a}_{k+1}} = \left( (t_1, l_1), (t_2, l_2), \ldots, (t_{N}, l_{N})\right)$, where $l_1=a_{k, l}$, $l_N = a_{k+1, l}$, $t_1 = a_{k+1, s} - T_W$ and $t_N = a_{k+1, s}$. The trajectory is generated by taking into consideration the following factors: (1) the straight distance between the start position and the end position, (2) a discomfort value to walk nearby furniture or walls, (3) a discomfort value for abrupt turning of the body, and (4) a difference between the step length and the preferable stride length \cite{Jiang2020Automaticspatialattribute}, while $T_W$ is determined by a walking speed after the trajectory is made. In the experiment, we assumed a fixed walking speed of 68.75 [cm/s].

Let $W_A$ be the successive trajectories according to an activity sequence $A$ as $W_A = (0, \bm{a}_{1, l}) + W_{\bm{a}_1, \bm{a}_2} + W_{\bm{a}_2, \bm{a}_3} + \cdots + W_{\bm{a}_{N_A-1}, \bm{a}_{N_A}}$. We also assumed that the resident does not move while doing a single activity, except for \anomaly{wandering}.

\subsubsection{Sensor data acquisition}
Sensor data are obtained from sensors attached to the rooms as the result of the resident's activity. In practice, sensors are activated by rules applied to an activity sequence and walking trajectories. Different activation rules are applied depending on the kind of sensors:
\begin{description}
\setlength{\itemsep}{1pt}
\setlength{\leftskip}{0.5cm}
\item[PIR sensors] are activated when the resident, whose body radius is 10 [cm], enters in their detection area of around radius 0.5 [m], and are inactivated when the resident leaves the area. Since PIR sensors are motion sensors, they are also inactivated if the resident stays still at a location without movement. Sampling rate is 10 [Hz].
\item[PR sensors] are activated when the resident stands on the mat with the sensors. Sampling rate is 10 [Hz].
\item[COST sensors] are constantly activated while the associated equipments are in use by the corresponded activities, such as a power sensor on TV is activated by activity \activity{watch TV} and a flow sensor on kitchen faucets by \activity{cooking}. Sampling rate is 1 [Hz]. 
\end{description}
All samplings start at the same time and are synchronized.

Those sensors are chosen because they are available at low cost, unintrusive, less physical burden and less privacy violation to the resident, that satisfy conditions preferred in behavioral sensing \cite{Pavel2013Theroleof}. The other sensors, e.g., video camera, microphone, radio frequency sensors, or wearable sensors used in some smart home \cite{Rashidi2012Asurveyon, Chan2008Areviewof} are excluded due to strong impact on one of them. Video cameras and microphones are unintrusive, but some elderly do not like to being watched or recorded. Video camera are also affected by light conditions. For wearable sensors, forcible attachment of them is unrealistic for a forgetful person with dementia. Therefore, we use above three kinds of binary-output sensors.

\subsection{Anomaly simulator}
\begin{figure}[tbp]
\centering
\includegraphics[width=\linewidth]{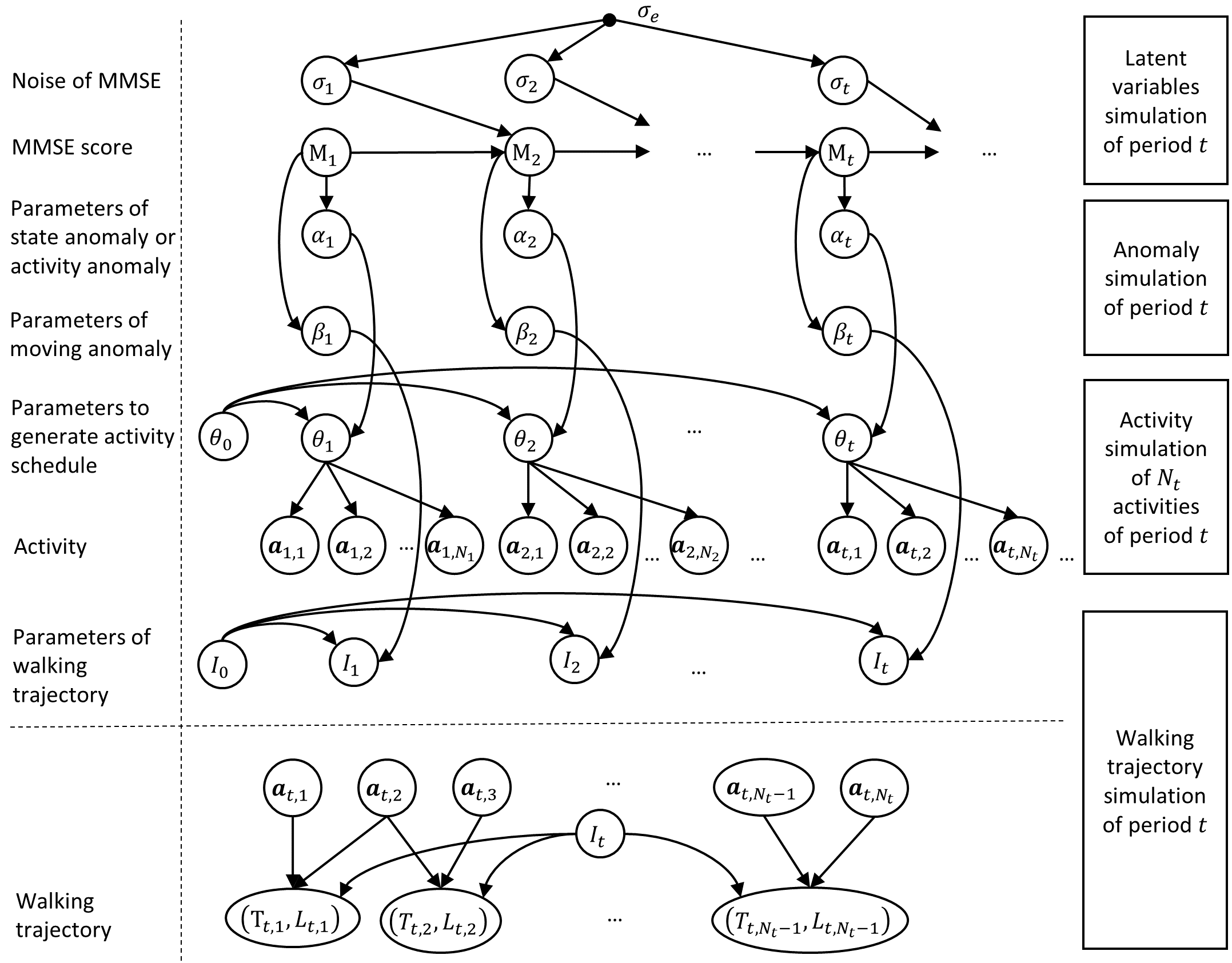}
\caption{A graphical model of anomalies. For example, MMSE score in $t$th month $M_t$ is generated according to the previous value $M_{t-1}$ and the noise $\sigma_{t}$. Random variables $\alpha_t$ and $\beta_t$ denote the parameters of state anomaly and activity anomaly, respectively (Table \ref{tab:six_implemented_anomalies}). They are conditioned by $M_t$ and condition the parameters ($\theta_t$ and $I_t$) of activities and walking trajectory, respectively.}
\label{fig:graphical_model}
\end{figure}

Last, we prepare an anomaly simulator to produce multiple anomalies. In this paper, we assume that anomalies are correlated to each other and governed in common by a single latent variable associated to the degree of dementia. We classify anomalies into three types:
\begin{description}
\setlength{\itemsep}{1pt}
\setlength{\leftskip}{0.5cm}
\item[State anomalyies] happen periodically. A state anomaly is observed in a change of the statistics. For example, the duration time of sleep will be increased month-to-month when the resident suffers from \anomaly{semi-bedridden}, and the frequency of an outing will be decreased for the resident being \anomaly{housebound}.
\item[Activity anomalies] are associated with a special activity. They are observed also in a change of the frequency/duration. \anomaly{Wandering} and \anomaly{forgetting (to turn off the stove)} are two examples.
\item[Moving anomalies] are observed as a strange way of walking. Falls are examples. When the resident falls, the walking speed becomes zero for some seconds or more.
\end{description}
The activity simulator and anomaly simulator are related statistically to each other as illustrated in a graphical model (Fig. \ref{fig:graphical_model}). In this paper, we realized six typical anomalies with a motion, which can be detected by PIR sensors, PR sensors or COST sensors (Table \ref{tab:six_implemented_anomalies}). 

In this study, to make the situation realistic, we focus on patients with mild cognitive impairment or with mild Alzheimer's disease, in their early stage of dementia with mini-mental state examination (MMSE) score of 19-27 \cite{Barry2010StagingDementia}. They are also supposed to be living alone. To quantize the degree of cognitive degeneration, we use a MMSE as a governing latent variable (\ref{sec:characteristics_of_anomaly}). Many anomalies are associated to MMSE in duration or in frequency \cite{Algase2009EmpiricalDerivationand, Algase2009Newparametersfor, Anstey2006An8-yearprospective}. 

\begin{table}
\centering
\caption{Implemented six anomalies (two of each type).}
\label{tab:six_implemented_anomalies}
\begin{tabular}[tbp]{p{1cm}p{3cm}p{10cm}}
\\
\hline
Type & Anomaly & Sympton\\ \hline
State & \anomaly{Semi-bedridden} & The resident tends to sleep and avoid to move. For example, we suppose that semi-bedridden has the anomaly status as the grade 2 or grade 3 in Eastern Cooperative Oncology Group (ECOG) performance status \cite{Oken1982Toxicityandresponse} (\ref{sec:ECOG}).\\
      & \anomaly{Housebound} & The resident is confined to their home more than one week 
      \cite{Ganguli1996Characteristicsofrural}.\\\hline
Activity & \anomaly{Wandering} & The resident walks around with anomalous travel patterns, such as back-and-forth repetition, circling repetition and randomly roundabout \cite{Martino-Saltzman1991TravelBehaviorof}.\\
         & \anomaly{Forgetting} & The resident forgets to turn off home appliances or faucets. In particular, high electric usage has been caused by forgetting to turn off the equipment.\\\hline
Moving & \anomaly{Fall while walking} & The resident suddenly falls down while he or she is walking \cite{Chaccour2016Fromfalldetection}.\\
       & \anomaly{Fall while standing} & The resident loses their balance when he or she starts standing up from sitting, or vice versa \cite{Nyberg1995Patientfallsin}.\\
\hline
\end{tabular}
\end{table}

\subsubsection{Implementation of MMSE}
MMSE can be considered as a time series. MMSE tends to rapid change in cases of severe dementia \cite{Barry2010StagingDementia} or advanced age \cite{Nagaratnam2022TrajectoriesofMini-mental}. Let $M_m$ be the random variable of MMSE score in $m$th [month]. In our model, sequence $\{M_m\}$ decreases with a constant $c$ and receives a random error $\sigma_m$ according to the following formula:
\begin{equation}\label{eq:MMSE}
M_{m+1} = M_m - c + \sigma_m,~~~c > 0, ~~~\sigma_m\sim \mathcal{N}(0,~\sigma_e^2).
\end{equation}
In this paper, we used $c = (29 - 19.5)/9 /12$, reducing from 29 to 19.5 over nine years, referring to the average cognitive degeneration in an early stage of dementia \cite{Barry2010StagingDementia}. A lower value of $M_m$ typically increases more the occurrence frequency of an anomaly and/or the duration time (see Table \ref{tab:frequency_and_duration_of_anomaly}).

\subsubsection{Frequency and duration of each anomaly}
\begin{figure}[tbp]
\centering
\includegraphics[width=\linewidth]{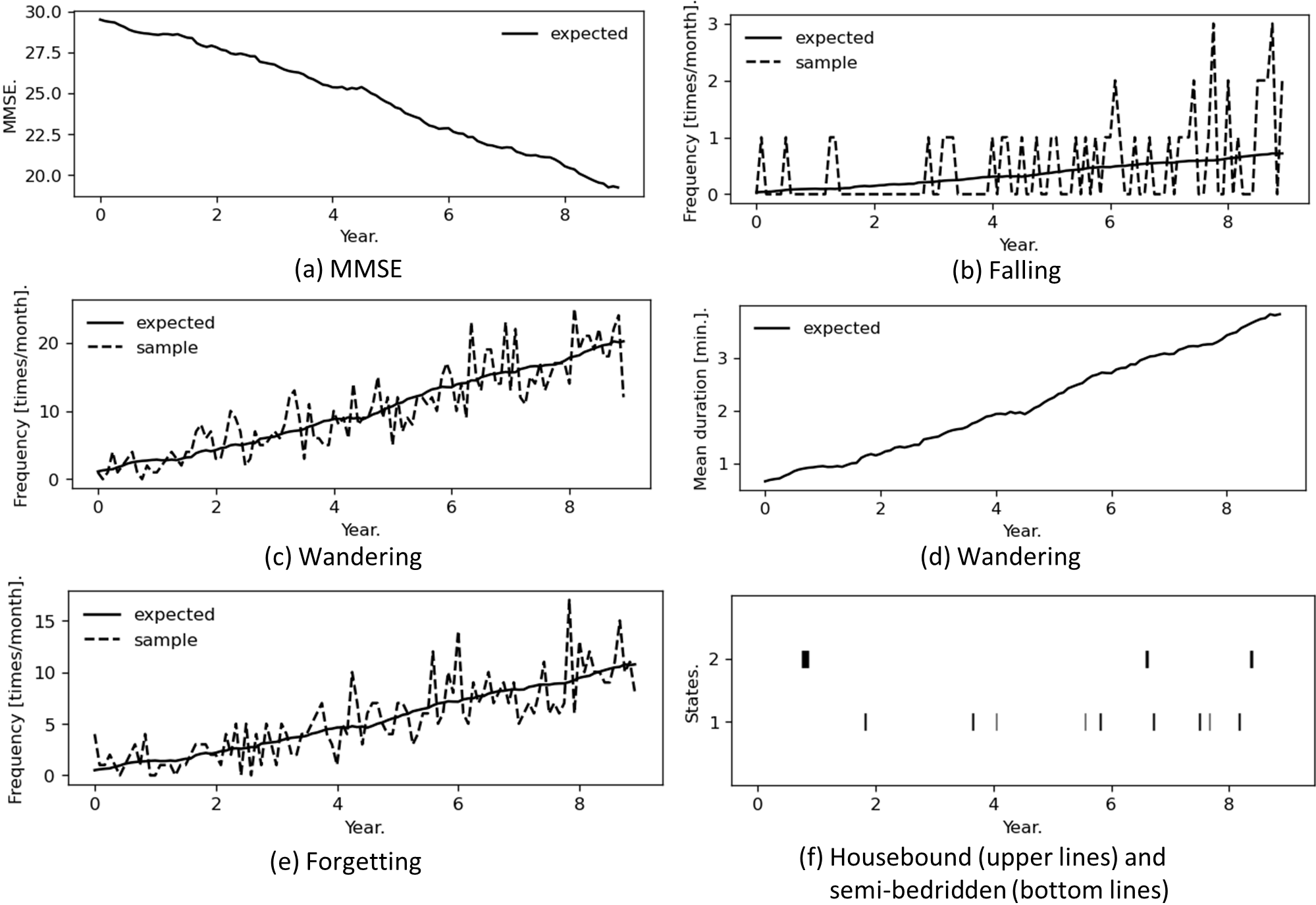}
\caption{Transitions of frequency and duration of anomalies over nine years (solid lines are the expected values and dashed lines are the sampled values). (a) Starting at 29, MMSE score decreases (the value is plotted at every 30 days). (b-e) The frequency or duration time of anomalies (\anomaly{wandering}, \anomaly{forgetting}, and \anomaly{fall}) increase as MMSE decreases. (f) \anomaly{Housebound} and \anomaly{semi-bedridden} happen according to the frequency and duration defined by Table \ref{tab:frequency_and_duration_of_anomaly}, independently of MMSE.}
\label{fig:anomaly_parameter_transitions}
\end{figure}

We assume that each anomaly $a$ occurs with mean frequency $F_{a, m}$ in the $m$th month and continue $D_{a, m}$ (the unit range changes from sec. to day) on average, where $a$ is one of \anomaly{semi-bedridden}, \anomaly{housebound}, \anomaly{wandering}, \anomaly{forgetting}, \anomaly{fall while walking}, and \anomaly{fall while standing}. Their values are set to a constant in the first two and change depending on $M_m$ in the others (Table \ref{tab:frequency_and_duration_of_anomaly}). An example is shown in Fig. \ref{fig:anomaly_parameter_transitions}.

\begin{table}
\centering
\caption{The mean frequency and duration of six anomalies. The mean $F_{*, m}$ of frequencies in $m$th month and the mean $D_{*, m}$ of durations are shown. Except for the first two, they are functions of MMSE score $M_m$.}
\label{tab:frequency_and_duration_of_anomaly}
\begin{tabular}[tbp]{p{2cm}p{3cm}p{3cm}p{5cm}}
\\
\hline
Type     & Anomaly                       & Frequency $F_{m}$ & Duration $D_{m}$\\ \hline
State    & \anomaly{Semi-bedridden}      & $1/20$            & $30$ [day]\\
         & \anomaly{Housebound}          & $1/10$            & $14$ [day]\\
Activity & \anomaly{Wandering}           & $-1.86 M_m + 56$  & $-0.31M_m + 9.8$ [min.]\\
         & \anomaly{Forgetting}          & $- M_m + 30$      & depend on an activity sequence\\
Moving   & \anomaly{Fall while walking}  & $- M_m/15 + 2$    & $30$ [sec.]\\
         & \anomaly{Fall while standing} & $- M_m/15 + 2$    & $30$ [sec.]\\
\hline
\end{tabular}
\end{table}

\subsubsection{Implementation of \anomaly{semi-bedridden} and \anomaly{housebound}}
The starting time of a state anomaly (\anomaly{semi-bedridden} or \anomaly{housebound}) is sampled according to a Poisson distribution with mean $F_{a, m}$, and the duration is sampled according to a normal distribution with mean $D_{a, m}$ and standard deviation $D_{a, m}/5$, independently of $M_m$.

The symptoms of state anomalies appear as a statistical change of some normal activities (Table \ref{tab:parameters_of_each_activity}). When the resident becomes \anomaly{housebound}, the times of \activity{go out} belonging to $S_N$ decreases by $\lambda_{\activity{go out} } = 1/14$ [times/day] and the duration is also reduced by $\mu_{\activity{go out} , d} = 20$ [min.]. In addition, the resident has less opportunities to call by $\lambda_{\activity{use the phone}} = 1/3$ [times/day] and $\mu_{\activity{use the phone}} = 10$ [min.]. Similarly, when the resident becomes \anomaly{semi-bedridden}, \activity{nap} and \activity{rest} increase and \activity{go out} decreases with their parameters as $\mu_{\activity{nap}, d} = 40$ [min.], $\mu_{\activity{rest}, d} = 60$ [min.] and $\lambda_{\activity{go out} } = 1/7$ [times/day], $\mu_{\activity{go out} , d} = 20$ [min.].

\subsubsection{Implementation of \anomaly{wandering}}
$\anomaly{Wandering}$ belongs to $S_N$ and has a mean duration $D_{\anomaly{wandering}, m}$, a mean frequency $F_{\anomaly{wandering}, m}/30$ and standard deviation $F_{\anomaly{wandering}, m}/150$, a day. Their values are determined by $M_m$ as shown in Table \ref{tab:frequency_and_duration_of_anomaly}. The resident suffering \anomaly{wandering} walks around the rooms with anomalous travel patterns. We implemented \anomaly{wandering} as the situation that the resident continues to walk to reach a goal point via several staging points randomly selected. The walking trajectory $W_{\bm{w}, \bm{a}} = ((w_{k, s}, w_{k, l}), \ldots, (t_x, l_1), \ldots, (t_y, l_n), \ldots, (a_{k+1, s}, a_{k+1, l}))$, on a transition from a \anomaly{wandering} $\bm{w}=(\anomaly{wandering}, w_{k, s}, w_{k, d}, w_{k, l})$ to an another activity $\bm{a}=(a_{k+1, n}, a_{k+1, s}, a_{k+1, d}, a_{k+1, l})$, is realized by randomly selected staging points $l_1, l_2, \ldots, l_n$.

\subsubsection{Implementation of \anomaly{forgetting}}
The frequency of \anomaly{forgetting} is sampled according to a Poisson distribution with mean $N_{\anomaly{forgetting}, m}/30$. The objective of \anomaly{forgetting} is randomly sampled from the activities that use some home appliances. If it happens, the resident \anomaly{forgets to turn off} the home appliance associated to it. The home appliance is safely turned off when the resident comes back to near the sensor location at next time. When the multiple home appliances are used in the activity, e.g., \activity{Cook} with the faucet and the kitchen stove, randomly selected one is left as turned on.

\subsubsection{Implementation of \anomaly{falls}}
The frequencies of \anomaly{falling while walking} and \anomaly{falling while standing} are sampled according to Poisson distribution with mean $F_{\anomaly{fall while walking}, m}/30$ and that with mean $F_{\anomaly{fall while standing}, m}/30$, respectively. The falling location is randomly sampled from the walking trajectory. If the resident \anomaly{falls while standing} on a trajectory $W=\{(t_1, l_1), \ldots, (t_N, l_N)\}$, the resident stops suddenly at the starting point $l_1$, and then restarts to walk after 30 [sec.]. For \anomaly{fall while walking}, the falling location is somewhere randomly chosen from $\{l_2, \ldots, l_{N-1}\}$ and the resident does not move for 30 [sec.]. During a fall, the body radius of a resident changes from 10 [cm] to $40$ [cm], representing their fall posture.

\section{Qualitative evaluation of activity seuqneces}\label{sec:qualitative_evaluation}
Sensor data are generated from the resident's activity sequence and the walking trajectories that have plausible reality as stated in \cite{Jiang2021SISG4HEI_Alpha:Alphaversion, Jiang2020Automaticspatialattribute}. So, we evaluate the reality of activity sequences, instead of the reality of sensor data.

\begin{figure}[tbp]
\centering
\includegraphics[width=160mm]{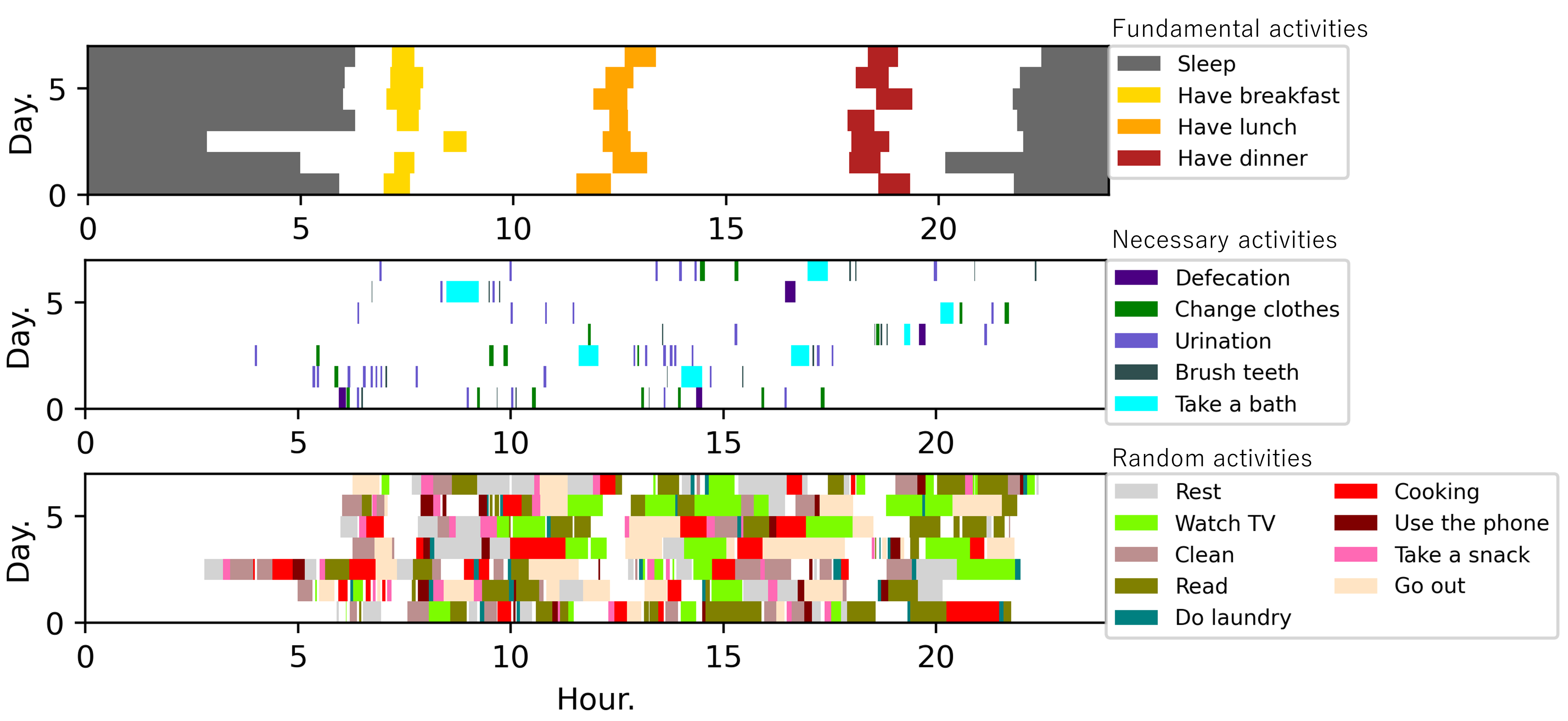}
\caption{Simulated activity sequence over 7 days (seven rows in each subfigure). Three types of activities are shown top to down: fundamental, necessary and random activities. There are 18 normal activities.}
\label{fig:activity_sequence}
\end{figure}

For qualitative evaluation, we investigate the validity of the generated activity sequence. The floor plan and sensor arrangement are shown in Fig. \ref{fig:floor_plan_and_sensor_arrangement}. The parameters of normal activities are shown in Table \ref{tab:parameters_of_each_activity}, and those of anomalies are shown in Table \ref{tab:frequency_and_duration_of_anomaly}. The walking speed is set to 68.75 [cm/s]. The value of MMSE score decreases almost linearly month to month (Fig. \ref{fig:anomaly_parameter_transitions}). Fig. \ref{fig:activity_sequence} shows the first week (seven days) of normal activities generated by the proposed simulator. We can confirm that 1) each of fundamental activities occurs at almost the same time, 2) each of necessary activities occurs several times in a day, and 3) random activities occur randomly but so as to fill remaining time. Although elderly tend to be punctual for meal and sleep, such a tendency is note observed in the simulation. This is because this simulation is a mixture of activities of many elderly people. It is easy to set each of some activities at the same time but we think of a variety of activities as more important in the simulator for anomaly detection.

\begin{figure}[tbp]
\centering
\includegraphics[width=140mm]{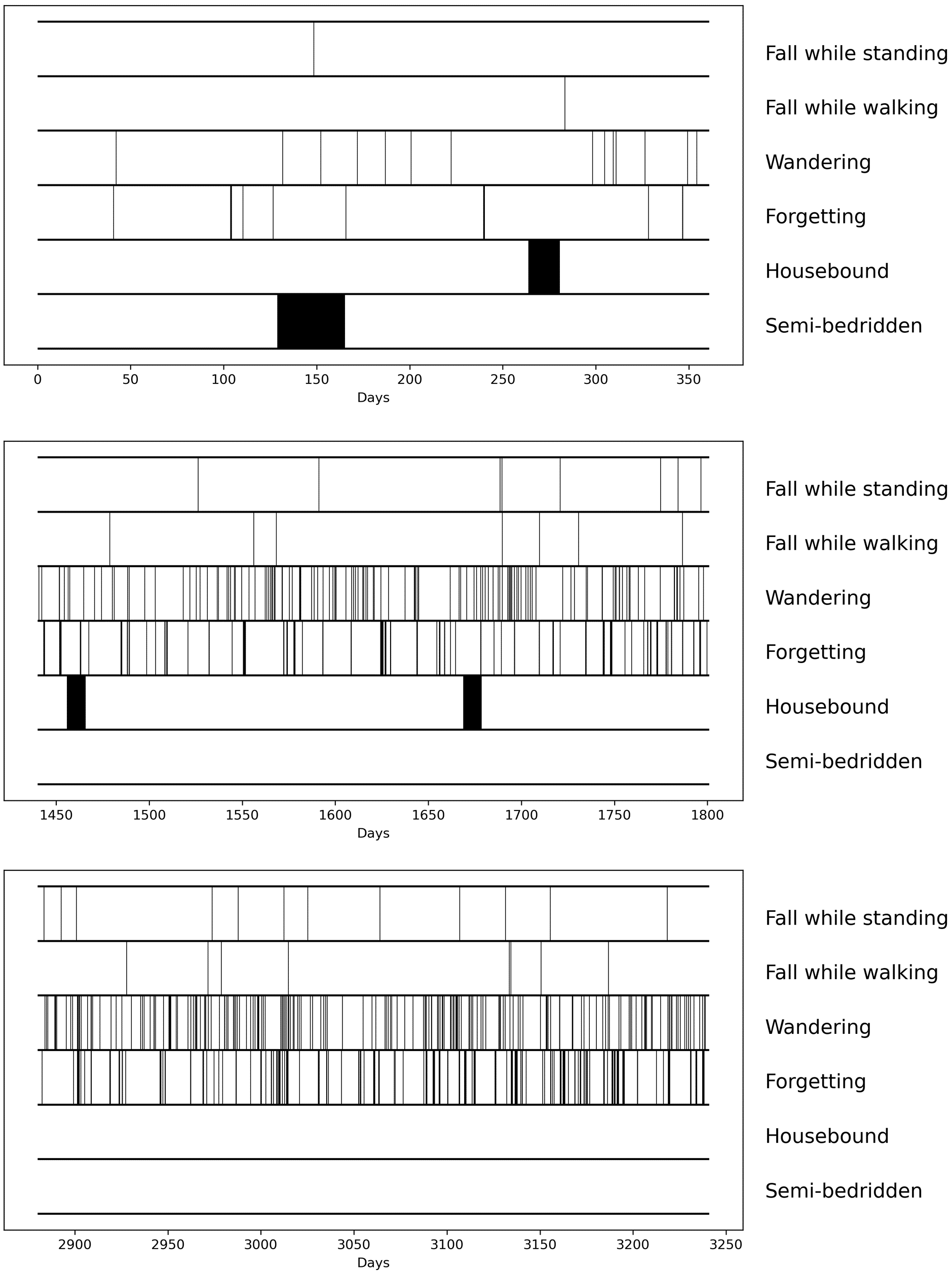}
\caption{Simulated six anomalies in three years (1st, 5th and 9th years) of simulated 9 years (old to new). The first four anomalies increase their frequencies according to the progress of dementia (MMSE). The last two anomalies are independent to MMSE (Table \ref{tab:frequency_and_duration_of_anomaly}).}
\label{fig:anomaly_occurrence}
\end{figure}

\begin{figure}[tbp]
\centering
\includegraphics[width=100mm]{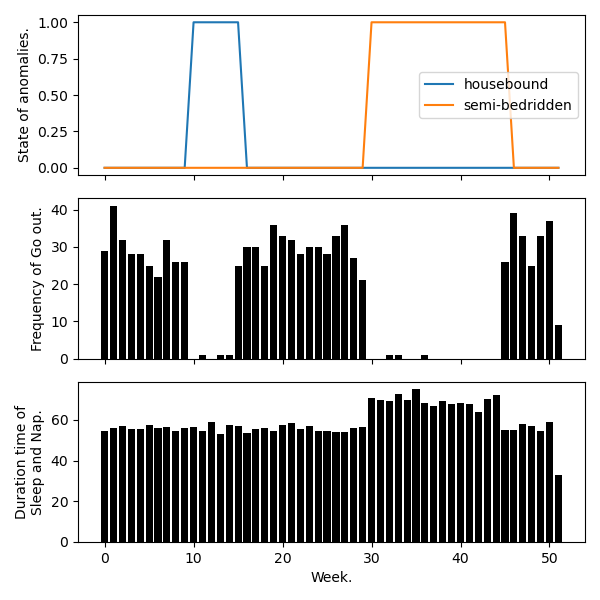}
\caption{Change of activities \activity{go out}, \activity{nap} and \activity{sleep} over 50 weeks. The occurrence is shown in the top figure, the frequency of \activity{go out} is shown in the second figure, and the duration of \activity{nap} and \activity{sleep} is shown in the third figure. Some decay of frequency and a little amount of increase of duration are observed according to two of anomalies.}
\label{fig:block_time_histogram}
\end{figure}

\begin{figure}[tbp]
\centering
\includegraphics[width=\linewidth]{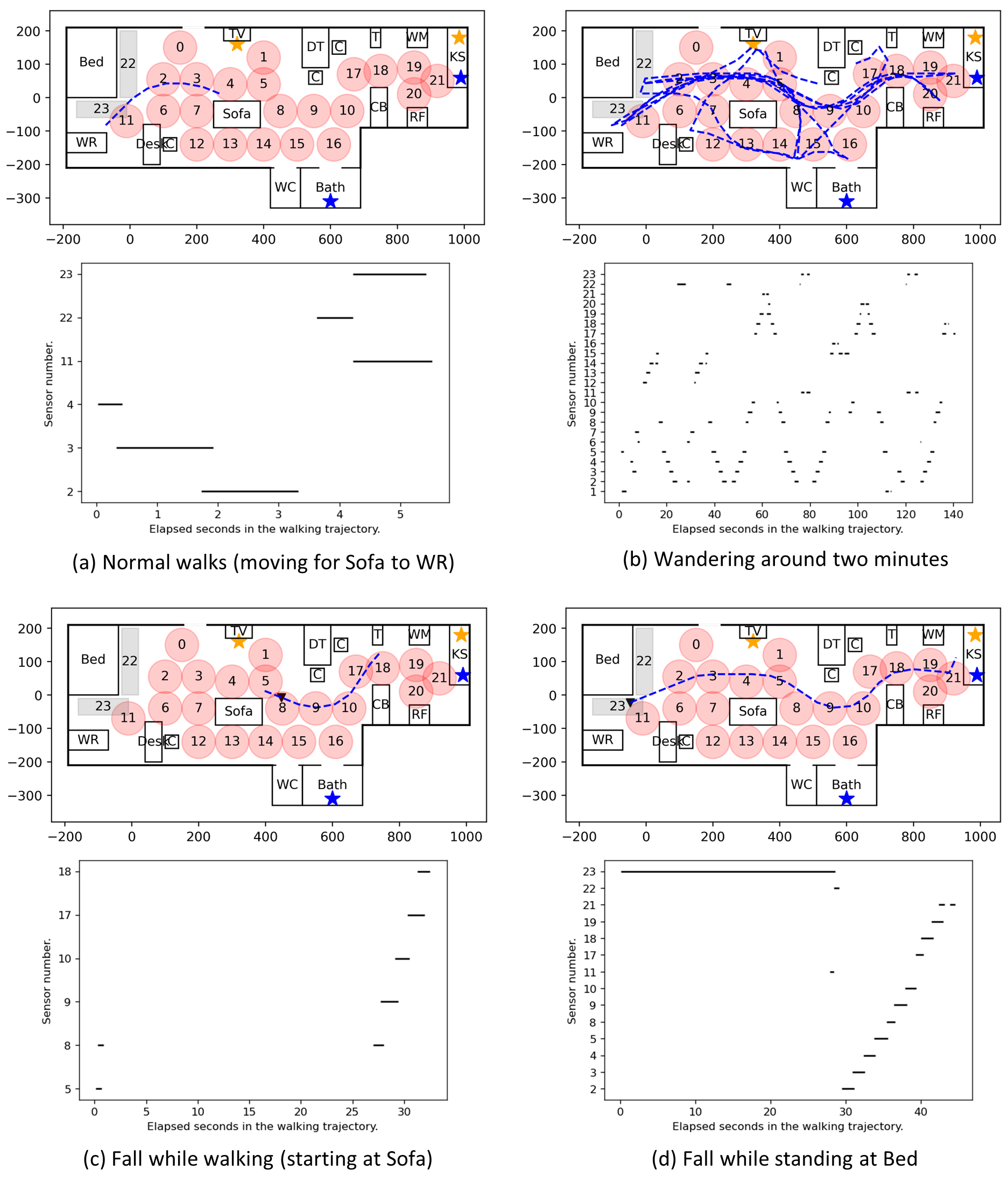}
\caption{Comparison of walking routes and sensor responses between a normal walk (a) and three anomalous walks (b), (c) and (d). Sensor numbers 1-21 are PIR and 22 and 23 are PR sensors. In (c) and (d), after a fall at some position (indicated by a black triangle), the resident restarts walking.}
\label{fig:sensor_activation_in_walks}
\end{figure}

Six anomalous patterns over nine years realized by the proposed simulator are shown in Fig. \ref{fig:anomaly_occurrence}. As shown in Fig. \ref{fig:anomaly_occurrence}, as time goes, the frequency and duration of the first four anomalies increase, with the decrease of MMSE. \anomaly{Housebound} and \anomaly{semi-bedridden} decrease the frequency of \activity{go out} and increase the duration time of \activity{nap} as shown in Fig. \ref{fig:block_time_histogram}. An increase of \anomaly{forgetting} imposes more cost of home appliances. \anomaly{Wandering}, \anomaly{fall while walking} and \anomaly{fall while standing} cause anomalous walking patterns, bringing strange patterns in sensor output (Fig. \ref{fig:sensor_activation_in_walks}). As shown in Fig. \ref{fig:sensor_activation_in_walks}, binary PIR sensors and PR sensors are promising to detect such strange patterns.

\section{Quantitative evaluation}\label{sec:quantitative_evaluation}
Next, we evaluate quantitatively the reality of activity sequences.

\subsection{Metrics of day-to-day similarity \cite{Kristiansen2018AnActivityRule}}
We compared the generated activities with real activity sequences \cite{Van2008Accurateactivityrecognition, Cook2013CASAS:ASmart}. To compare those activities, we sampled activities at frequency $f$ [Hz] and represented them as a character sequence of a constant length. For example, at $f=2$ [Hz], 20-minute \activity{have lunch} (L) following 30-minute \activity{nap} (N) is transformed to LL$\cdots$LNN$\cdots$N (2400 times L and 3600 times N).

Given two daily activity (character) sequences of $\ActSequences{A}$ and $\ActSequences{B}$, we measure the similarity $\textrm{sim}(\ActSequences{A}, \ActSequences{B})$ by Levenshtein distance \cite{Levenshtein1965Binarycodescapable} as seen in \cite{Kristiansen2018AnActivityRule}. Two sequences over different days, $\ActSequences{A} = (A_{1}, A_{2}, \ldots, A_{m})$ and $\ActSequences{B} = (B_{1}, B_{2}, \ldots, B_{n})$ are compared as;
\begin{equation}
\rho(\ActSequences{A}, \ActSequences{B}) = \sum_{i}^{m}\sum_{j}^{n}\textrm{sim}(A_{i}, B_{j}) / mn,
\end{equation}
\begin{equation}
\rho(\ActSequences{A}, \ActSequences{A}) = \sum_{i}^{m}\sum_{j\neq i}^{m}\textrm{sim}(A_{i}, A_{j}) / \binom{m}{2},
\end{equation}
\begin{equation}
\delta_{intra}(\ActSequences{A}, \ActSequences{A}^{*}) = \lvert \rho(\ActSequences{A}, \ActSequences{A}) - \rho(\ActSequences{A^{*}}, \ActSequences{A^{*}})\rvert,
\end{equation}
\begin{equation}
\delta_{inter}(\ActSequences{A}, \ActSequences{A}^{*}) = \lvert \rho(\ActSequences{A}, \ActSequences{A^{*}}) - \rho(\ActSequences{A^{*}}, \ActSequences{A^{*}})\rvert,
\end{equation}
where $\ActSequences{A}^{*}$ means a real activity sequence.
From $\rho(\ActSequences{A}, \ActSequences{A})$, we can know the variety (or inversely, self-similarity) of daily schedules within $m$ days of $\ActSequences{A}$, that is, a large value of $\rho(\ActSequences{A}, \ActSequences{A})$ implies that there is a large variety over days. While, from $\rho(\ActSequences{A}, \ActSequences{B})$, we can know the similarity between $m$ days of $\ActSequences{A}$ and $n$ days of $\ActSequences{B}$. From $\delta_{intra}(\ActSequences{A}, \ActSequences{A}^{*})$, we can know how $\ActSequences{A}$ differs from $\ActSequences{A^{*}}$ in variety. From $\delta_{inter}(\ActSequences{A}, \ActSequences{A}^{*})$, we can know how relatively $\ActSequences{A}$ differs from $\ActSequences{A^{*}}$ in similarity. A small value of $\delta_{intra}(\ActSequences{A}, \ActSequences{A}^{*})$ means that $\ActSequences{A}$ holds a similar variety, while a small value of $\delta_{inter}(\ActSequences{A}, \ActSequences{A}^{*})$ means that $\ActSequences{A}$ and $\ActSequences{A^{*}}$ is similar in contents. 

\subsection{Dataset description}
We used two real activity sequences extracted from two datasets (\cite{Van2008Accurateactivityrecognition} and \cite{Cook2013CASAS:ASmart}).

\subsubsection{Kasteren dataset \cite{Van2008Accurateactivityrecognition}}\label{sec:Kasteren}
We compared five activity sequences generated in the following five ways. For the detail, see \ref{sec:parameter_fitting}. 
\begin{description}
\setlength{\itemsep}{1pt}
\setlength{\leftskip}{0.5cm}
\item[$A_{Kasteren}$ (24 days)] The real activity sequence for 24 days of a 26 year-old resident living alone \cite{Van2008Accurateactivityrecognition}. Some preprocessing and outlier removal were applied (\ref{sec:preprocessing_of_real_data}). The original data consist of seven activities of \activity{prepare breakfast}, \activity{take shower}, \activity{prepare dinner}, \activity{go to bed}, \activity{use toilet}, \activity{leave house}, and \activity{get drink}. We added one more new activity, \activity{other} (to fill the blank time). 
\item[$A_{Proposed}$ (24 days)] An activity sequence for 24 days generated by our activity simulator. The parameter values were learned from $A_{Kasteren}$ in the way shown in \ref{sec:parameter_fitting}. We regarded four activities (\activity{take breakfast}, \activity{take shower}, \activity{take dinner} and \activity{go to bed}) as fundamental activities, and one activity (\activity{use toilet}) as necessary activity, and three activities (\activity{go out}, \activity{get drink}, and \activity{other}) as random activities. Standard deviations of start time and duration time of \activity{go to bed} are shrunk to one-third to avoid overlap of fundamental activities.
\item[$A_{RanD}$ (24 days)] An activity sequence for 24 days generated at random whose activities have random duration according to normal distributions learned from $A_{Kasteren}$. The activities were generated in a chronological order, and a next activity were randomly and uniformly selected from seven activities above. This data becomes a baseline data.
\item[$A_{RanS}$ (24 days)] An activity sequence for 24 days generated in a similar way with $A_{RanD}$, but the sleep is well controlled in its starting time in according with $A_{Kasteren}$. The sleep activity is determined first and then other activities follow. This modification is introduced because sleep time occupies a large part of a day. This data is another baseline data.
\item[$A_{ADLSim}$ (24 days)] The explicit activity sequence is not given, but some values of metrics are given by \cite{Kristiansen2018AnActivityRule} as Sim2. They used probability distributions to select next activities with rules that prioritize activity among activities.
\end{description}

\subsubsection{CASAS Aruba dataset \cite{Cook2013CASAS:ASmart}}
We used one more dataset called CASAS Aruba \cite{Cook2013CASAS:ASmart}. We compared six activity sequences generated as follow. This data is about nine times longer than Kasteren dataset and the resident is older than that of Kasteren dataset, so we divided it into training sequence and testing sequence at 2:1.
\begin{description}
\setlength{\itemsep}{1pt}
\setlength{\leftskip}{0.5cm}
\item[$A_{Aruba}$ (220 days)] The real activity sequence for 220 days of an elderly woman living alone, to whom her grandchildren and children visit regularly \cite{Cook2013CASAS:ASmart}. Preprocessing was executed as explained in \ref{sec:preprocessing_of_real_data}. The original data consist of eleven activities of \activity{meal preparation}, \activity{relax}, \activity{eating}, \activity{work}, \activity{sleeping}, \activity{wash dishes}, \activity{bed to toilet}, \activity{enter home}, \activity{leave home}, \activity{housekeeping} and \activity{resperate}. As a preprocessing, we deleted short and intermittent activities and adopted six activities, and added five of \activity{take breakfast}, \activity{take lunch}, \activity{take dinner}, \activity{go out}, \activity{nap} and \activity{other}. 
\item[$A_{ArubaTrain}$ (146 days)] First 146 days of $A_{Aruba}$ were selected as training data. $A_{ArubaTrain}$ is used to fit models in simulators.
\item[$A_{ArubaTest}$ (74 days)] Last 74 days of $A_{Aruba}$ were selected as test data. $A_{ArubaTest}$ is used to measure the test error.
\item[$A_{Proposed}$ (74 days)] An activity sequence for 74 days generated by the proposed activity simulator. The parameter values were learned from $A_{ArubaTrain}$ as explained in \ref{sec:parameter_fitting}. We regarded four activities (\activity{take breakfast}, \activity{take lunch}, \activity{take dinner} and \activity{sleeping}) as fundamental activities, four activities (\activity{nap}, \activity{wash dishes}, \activity{meal preparation} and \activity{housekeeping}) as necessary activities, and remaining activities as random activities. Standard deviations of the start time and duration time of \activity{sleeping} were shrunk to one-half to avoid an overlapping of fundamental activities.
\item[$A_{RanD}$ (74 days)] Data generated with $A_{ArubaTrain}$ in a similar way used for $A_{RanD}$ of Kasteren dataset.
\item[$A_{RanS}$ (74 days)] Data generated with $A_{ArubaTrain}$ in a similar way used for $A_{RanS}$ of Kasteren dataset.
\end{description}

\subsection{Result}
The results on Kasteren dataset are shown in Table \ref{tab:result_of_Kasteren}. In Table \ref{tab:result_of_Kasteren}, we see that $A_{Proposed}$ and $A_{ADLSim}$ have almost the same degree of variety as $A_{Kasteren}$, from the values of $\delta_{intra}$, and $A_{RanS}$ is the most similar to the real ones, from the values of $\delta_{inter}$, and $A_{Proposed}$ is the second. This is because $A_{RanS}$ borrows the schedule of \activity{sleeping} from the real data.

The results on CASAS Aruba dataset are shown in Table \ref{tab:result_of_Aruba}. In Table \ref{tab:result_of_Aruba}, $A_{Proposed}$ follows $A_{ArubaTrain}$ both in $\delta_{intra}$ and $\delta_{inter}$, and is better than two baselines in $\delta_{intra}$. That is, $A_{Proposed}$ has the same degree of variety in activity with actual activities a day and is more similar than the two random assignments. Scheduling of \activity{sleeping} is still important but its sole scheduling is insufficient to realize real activities. Note that, because anomalies are almost out of consideration, short activities are almost ignored in this measure.

\begin{table}
\centering
\caption{Similarity between activity sequences of Kasteren dataset and simulation. All values are the mean for 10 trials (the standard deviation). $\rho (A, A)$ means the intra-difference in the activity sequence in $A$, while $\rho (A_1, A_2)$ means the inter-difference between $A_1$ and $A_2$. Lower values of $\delta_{intra}(A, A_{Kasteren})$ and $\delta_{inter}(A, A_{Kasteren})$ mean that $A$ is closer to $A_{Kasteren}$ in variety and similarity, respectively. For reference, $\rho (A_{ADLSim}, A_{ADLSim})$ is taken from \cite{Kristiansen2018AnActivityRule}, while $\rho (A_{Kasteren}, A_{ADLSim})$ was not mentioned.}
\label{tab:result_of_Kasteren}
\begin{tabular}[tbp]{p{1.8cm}p{3.3cm}p{3.3cm}p{3.3cm}p{3.3cm}}
\\\hline
data ($A$) & $\rho (A, A) $ & $\rho (A, A_{Kasteren})$ & $\delta_{intra}(A, A_{Kasteren})^{*}$ & $\delta_{inter}(A, A_{Kasteren})^{**}$\\ \hline
$A_{Kasteren}$ & 1027.82 & - & 0 & -\\
$A_{Proposed}$ & ~~923.12 $(\pm 31.30)$ & 1262.27 $(\pm ~~35.36)$ & 104.70 $(\pm 31.30)$ & 234.46 $(\pm ~~35.36)$\\
$A_{RanD}$ & 1568.00 $(\pm 45.86)$ & 1676.81 $(\pm 106.81)$ & 540.18 $(\pm 45.86)$ & 649.00 $(\pm 106.81)$\\
$A_{RanS}$ & ~~799.95 $(\pm 76.83)$ & 1004.51 $(\pm ~~46.80)$ & 227.86 $(\pm 76.83)$ & ~~23.31 $(\pm ~~46.80)$\\
$A_{ADLSim}$ & 1130.96 $(\pm 90.47)$ & N.A. & 103.14 $(\pm 90.47)$ & N.A.\\
\hline
\end{tabular}\\
\flushleft{\small{
* $\delta_{intra}(A,A_{Kasteren}) = \lvert\rho (A, A)- \rho (A_{Kasteren}, A_{Kasteren})\rvert$.\\
** $\delta_{inter}(A, A_{Kasteren}) = \lvert\rho (A, A_{Kasteren}) - \rho (A_{Kasteren}, A_{Kasteren})\rvert$.
}}
\end{table}

\begin{table}
\centering
\caption{Similarity between activity sequences of CASAS Aruba dataset and simulation. All values are the mean for 10 trials (the standard deviation). $\rho (A, A)$ means the intra-difference in the activity sequence in $A$, while $\rho (A_1, A_2)$ means the inter-difference between $A_1$ and $A_2$. Lower values of $\delta_{intra}(A, A_{ArubaTest})$ and $\delta_{inter}(A, A_{ArubaTest})$ mean that $A$ is closer to $A_{ArubaTest}$ in variety and similarity, respectively.}
\label{tab:result_of_Aruba}
\begin{tabular}[tbp]{p{1.8cm}p{3.3cm}p{3.3cm}p{3.3cm}p{3.3cm}}
\hline
data ($A$) & $\rho (A, A)$ & $\rho (A, A_{ArubaTest})$ & $\delta_{intra}(A, A_{ArubaTest})^{*}$ & $\delta_{inter}(A, A_{ArubaTest})^{**}$\\ \hline
$A_{ArubaTest}$  & 1281.85 & - & 0 & -\\
$A_{ArubaTrain}$ & 1264.08 & 1329.98 & ~~17.78 & ~~48.13\\
$A_{Proposed}$   & 1309.92 $(\pm 19.63)$ & 1484.99 $(\pm 10.46)$ & ~~28.07 $(\pm 19.63)$ & 203.14 $(\pm 10.46)$\\
$A_{RanD}$       & 1961.20 $(\pm 58.69)$ & 2243.97 $(\pm 36.90)$ & 679.34 $(\pm 58.69)$ & 962.11 $(\pm 36.90)$\\
$A_{RanS}$       & 1801.87 $(\pm 24.45)$ & 1878.31 $(\pm 19.78)$ & 520.02 $(\pm 24.45)$ & 596.46 $(\pm 19.78)$\\
\hline
\end{tabular}\\
\flushleft{\small{
* $\delta_{intra}(A, A_{ArubaTest}) = \lvert\rho (A, A) - \rho (A_{ArubaTest}, A_{ArubaTest})\rvert$.\\
** $\delta_{inter}(A, A_{ArubaTest}) = \lvert\rho (A, A_{ArubaTest}) - \rho (A_{ArubaTest}, A_{ArubaTest})\rvert$.}
}
\end{table}

Comparison with the other agent scenario generation based activity simulators is unfortunately difficult for following reasons. The simulator \cite{Francillette2020Modelingthebehavior} focuses on only a few days and cannnot be applied easily to years. The simulators \cite{Bouchard2010SIMACT:A3D, Kristiansen2016Smoothandcrispy} need scripts or rules that are difficult to be learned from data, so that they are inferior in reality. The simulator \cite{Jiang2021SISG4HEI_Alpha:Alphaversion} using motivation values does not provide the detail of implementation of adjustment, so that we cannot use it accurately.

\section{Time consumption for simulation}

Execution time of simulation is measured. Our simulator is coded in Python on a computer with an AMD Ryzen 7 3700X 8-Core Processor 3.60 GHz and 32.0 GB RAM. Simulations to generate sensor data of 100 days are repeated 10 times. The average execution time is $166.2~(\pm 3.1)$ [s]. It corresponds to 100 [min.] for 10 years. This is almost the same as the simulator in \cite{Jiang2021SISG4HEI_Alpha:Alphaversion} because the both framework have the bottleneck for computing walking trajectories.

\section{Discussion}

\subsection{Reality in scenario and in anormalies}
Reality of simulated behaviors including anomalous ones and that of sensor data should be given the first priority to discuss the validity of the proposed simulator. However, unfortunately, comparison to real behaviors and sensor data are difficult because those are not available. At least they are short in the number and variety. It is not realistic to ask some elderly persons to live in a smart home for years. To make matters worse, real anomalies such as falls or housebound are uncertain, and even danger to simulate intentionally.

Usage of statistics is more safe and promising to keep reality. Such statistics on normal or anomalous activities are obtained more easily compared with a whole sensor data. This is the way that we took in this study. A hierarchical probabilistic model (Fig. \ref{fig:graphical_model}) is further useful to exploit the relationship between statistic parameters. If we need to specify a resident, it is sufficient to take statistics from the resident.

\subsection{Limitation \& Extendability}
The simulation approach has a fundamental limitation other than less reality. The current simulator might not be fully useful as it is. However, the simulator can be extended to several directions as below.

\begin{description}
\item[Activities] A long term habitual change can be implemented by making the related parameters depend on days. For examples, we may change the place to sleep by seasons, and decrease the times of cooking gradually. This simulator also limits the granularity of activities at level of body movement, avoiding finer movements including hands or legs, taking into the complexity. At expense of more complexity, we can make finer the granularity.
\item[Anomalies] More kinds of anomalies are easily dealt with if they are classified into one of \textit{state}, \textit{activity}, and \textit{moving} anomalies.
\item[Sensors] Other kinds of sensors such as a microphone and a thermometer can be added in the simulator. For this goal, however, it needs more detailed simulators such as a sound waves simulator and a body temperature simulator. For example, using those simulators, a cough might be detected by a microphone and the increase of the heart rate or a fever might be detected by a thermometer.
\item[Personalization] Adaptation to a specific person is made by tuning the parameters so as to meet the statistics of the person, e.g., time of breakfast, times of eats and so on.
\end{description}

In summary, the proposed simulator is a prototype of sophisticated data simulators. Even so, this simulator contributes already to the development of detection algorithms for typical anomalies. Actually, we notice that an existing benchmark classifiers \cite{Van2011Humanactivityrecognition} sometimes fails in detecting anomalies generated form the simulator.

\section{Conclusion}
In this paper, we have extended a state-of-the-art sensor data simulator \cite{Jiang2021SISG4HEI_Alpha:Alphaversion}, that treats wandering only, so as to produce anomalous patterns corresponding to six typical anomalies that can happen for elderly living alone: \anomaly{housebound}, \anomaly{semi-bedridden}, \anomaly{wandering}, \anomaly{forgetting}, \anomaly{fall while walking}, and \anomaly{fall while standing}. Their occurrences were statistically controlled mainly in frequency and in duration, and are governed by a single latent variable, MMSE score, an indicator of the degree of dementia. The proposed simulator is superior to previous simulators in the following: 1) the worsening over age are naturally realized, and 2) the correlation among anomalies are realized through MMSE score. The proposed method showed a high degree of similarity with real data in the activity sequence in two real datasets. The program code is available in public\footnote{under preparation}.

Future works include refining the model, e.g., adopting more flexible floor plans including another floor or yard, introducing habitual behaviors of the resident, and implementing 3D information for a few richer sensors. We plan to advance to development of anomaly detection algorithms on the basis of simulated data generated by the proposed method. In addition, with the algorithm, it is important to consider necessary number and kinds of sensors for anomalies under consideration.

\section*{Acknowledgments}
This work was partially supported by JSPS KAKENHI Grant Number 19H04128.

\bibliographystyle{IEEEtran}  
\bibliography{mybib.bib}  

\clearpage
\appendix

\section{Generation of a day schedule}\label{sec:algorithm_of_activity_scheduling}
We generate activity sequence in Algorithm \ref{alg:fill_T}.

\begin{figure}[!t]
\begin{algorithm}[H]
\caption{Generation of a daily schedule.}
\label{alg:fill_T}
\begin{algorithmic}[1]
\State Input: $S_F$, $S_N$ and $S_R$ that are the sets of fundamental activities, necessary activities and random activities.
\State Output: $A$ that is a sequence of activities of the day.
\State Let $N$, $Poi$, $U$ and $P_R$ be a normal distribution, a Poisson distribution, a uniform distribution, and a discrete distribution of random activities.
\State Let $\epsilon$ be an allowable minimum duration time, e.g., 30 seconds.
\Statex
\Procedure {behavioral\_simulation}{$S_F$, $S_N$, $S_R$}
\State $T\leftarrow [0:00, 24:00)$  // Time range of a day.
\State $A\leftarrow \emptyset$
\For {$a\in S_F$}  // Set up $S_F$.
\Repeat
\State $s\sim N(\mu_{a, s}, \sigma_{a, s}^2)$
\State $d\sim N(\mu_{a, d}, \sigma_{a, d}^2)$
\Until{UPDATE($T$, $A$, $a$, $s$, $d$) == Success}
\EndFor
\ForAll {$a\in S_N$}  // Set up $S_N$.
\State $b\sim Poi(\lambda_{a})$
\ForAll {$i \in \{1, 2,\ldots, b\}$}
\Repeat
\State $s\sim U(T)$
\State $d\sim N(\mu_{a, d}, \sigma_{a, d}^2)$
\Until{UPDATE($T$, $A$, $a$, $s$, $d$) == Success}
\EndFor
\EndFor
\While{$T\not= \emptyset$}  // Set up $S_R$.
\Repeat
\State $a\sim P_R(S_R)$
\State $s\leftarrow \min(T)$
\State $d\sim N(\mu_{a, d}, \sigma_{a, d}^2)$
\Until{UPDATE($T$, $A$, $a$, $s$, $d$) == Success}
\EndWhile
\State Return $A$
\EndProcedure
\Statex
\Procedure {Update}{$T$, $A$, $a$, $s$, $d$}
\If{$d<\epsilon$ $\lor$ ($\min\{b-a\mid [a, b)\subseteq T\}<\epsilon$) $\lor$ (the order of F in $A$ is invalid)}
\State return Failure
\Else  // When activities have enough duration.
\State $d^{'}\leftarrow\min(d, \max(\{ e\mid [s, s + e)\subseteq T\}))$
\State $T\leftarrow T\setminus [s, s + d^{'})$
\State $A\leftarrow A \cup (a, s, d^{'})$  // a is done for $d^{'}$ [m] from s.
\State return Success
\EndIf
\EndProcedure
\end{algorithmic}
\end{algorithm}
\end{figure}

\clearpage

\section{Characteristics of anomalies}\label{sec:characteristics_of_anomaly}
\subsection{ECOG performance status \cite{Oken1982Toxicityandresponse}}\label{sec:ECOG}
The eastern cooperative oncology group (ECOG) score is used to classify patients into 6 levels on its degree of independence (from 0 (fully active), to 5 (dead)). In particular, grade 2 means ''Ambulatory and capable of all self care but unable to carry out any work activities; Up and about more than 50\% of waking hours", and grade 3 means ''Capable of only limited selfcare; confined to bed or chair more than 50\% of waking hours." We regard grade 2 and 3 as symptoms of semi-bedridden.

\subsection{MMSE degree \cite{Folstein1975Mini-mentalstate}}
The mini-mental state examination (MMSE) degree \cite{Folstein1975Mini-mentalstate} is a numerical evaluation scale of dementia and is obtained from simple questions (e.g., what is the date today?). The lower points indicates the severe level of dementia. MMSE tends to rapid change in cases of severe dementia \cite{Barry2010StagingDementia} or advanced age \cite{Nagaratnam2022TrajectoriesofMini-mental}. The MMSE score tends to decrease by age \cite{Crum1993Population-basednormsfor}, and decreases gradually in the initial stage of dementia \cite{Barry2010StagingDementia}. So, we assume that the decrease can be approximated linearly.

\subsection{Relationship of anomalies with cognitive degeneration}
How cognitive degeneration is related to anomalies, is summarized in Table \ref{tab:relationship_of_anomalies}.

\begin{table}
\centering
\caption{Causes of anomalies.}\label{tab:relationship_of_anomalies}
\begin{tabular}[tbp]{p{3cm}p{12cm}}
\hline
Anomaly & causes\\ \hline
Semi-bedridden & Bedridden is preceded by various kinds of disabilities due to health problem such as stroke, weakened body by aging and injuries caused by falls \cite{Normala2020BedriddenElderly_Factors}. In addition, housebound is considered as a risk of bedridden \cite{Ishikawa2006Factorsrelatedto}.\\
Housebound & There is a correlation between the occurrence of houseboundness and dementia, but causal relationship is unknown \cite{Lindesay1993Houseboundelderlypeople}.\\
Wandering & MMSE score is known to be effective to connect the patient to the degree of wandering \cite{Algase2009EmpiricalDerivationand}. Especially, the duration time is reported to be significantly associated with the MMSE score \cite{Algase2009Newparametersfor}.\\
Forgetting & Age and frequency of forgetfulness are significantly correlated \cite{Szabo2011Cardiorespiratoryfitness_hippocampal}.\\
Fall while walking, Fall while standing & The longitudinal associations between MMSE and the falling risk are significantly strong at the individual level, and cognitive degeneration are predictive of falling risk\cite{Anstey2006An8-yearprospective}.\\
\hline
\end{tabular}
\end{table}

\section{Preprocessing of real data in the experiment}\label{sec:preprocessing_of_real_data}

Preprocessing on two real data of Kasteren dataset \cite{Van2008Accurateactivityrecognition} and CASAS Aruba dataset \cite{Cook2013CASAS:ASmart} are as follows.

\subsection{Kasteren dataset \cite{Van2008Accurateactivityrecognition}}

\begin{description}
\setlength{\itemsep}{1pt}
\setlength{\leftskip}{0.5cm}
\item[S1] The data of 24 days from 0:00 on February 26, 2008 to 24:00 on March 20, 2008 were extracted.
\item[S2] The following data was regarded as outliers and removed (1) \activity{leave house} longer than 23 hours (3 cases) (2) \activity{go to bed} less than 6 hours (4 cases), and (3) \activity{get drink} before 12 a.m. (4 cases).
\end{description}

\subsection{Aruba dataset \cite{Cook2013CASAS:ASmart}}

\begin{description}
\setlength{\itemsep}{1pt}
\setlength{\leftskip}{0.5cm}
\item[S1] The data for 220 days from November 4, 2010 to June 12, 2011 were used.
\item[S2] Overlapping between activities was removed such that one activity follows another just in time.
\item[S3] Duplicate data was removed.
\item[S4] A new activity \activity{other} was assigned to the vacancy in activities. In case that the duration is less than five minutes, then preceding activity was extended.
\item[S5] \activity{leave home} and \activity{enter home} labeled in a few seconds near going out were removed. Instead, \activity{go out} is assigned to the time.
\item[S6] \activity{eating} was separated into three of them, such as \activity{take breakfast} from 4:00 to 11:00, \activity{take lunch} from 11:00 to 16:00, \activity{take dinner} from 0:00 to 1:00, or 16:00 to 24:00.
\item[S7] \activity{sleeping} in a day time from 6 to 17 o'clock was replaced with a new label \activity{nap}.
\item[S8] \activity{bed to toilet} and \activity{respirate} were replaced with the preceding activity.
\end{description}

\section{Parameter setting of the activity simulator}\label{sec:parameter_fitting}

The parameters of activity occurrence are estimated from Kasteren dataset and CASAS Aruba dataset. The concrete values are shown in Table \ref{tab:parameters_fitted_by_Kasteren} and Table \ref{tab:parameters_fitted_by_Aruba}. Note that the average start time was calculated by excluding day, otherwise, for example, the average of 1 [d] 23 [h] and 3 [d] 0 [h] becomes 2 [d] 11 [h]. This problem was solved by using circular mean as \cite{Kristiansen2018AnActivityRule}.

Some parameters need special treatments. In Kasteren dataset, sampled \activity{go to bed} are sometime long and overlap with \activity{prepare breakfast}. To avoid such as overlap, the standard deviations of start time and duration time of \activity{prepare breakfast} are multiplied by $1/3$. In CASAS Aruba dataset, we did the same processing by halving ones of \activity{sleeping}.

\begin{table}
\centering
\caption{Parameters of Kasteren dataset in three types of activities (F: Fundamental, N: Necessary, R: Random). Here, $\mu_{a, s}$ and $\sigma_{a, s}$ denote the mean and the standard deviation of start time, $\mu_{a, d}$ and $\sigma_{a, d}$ denote those of duration time, and $\lambda_{a}$ is the mean frequency in a day, $p_{a}$ is an occurrence probability among random activities. The unit is a minute.}
\label{tab:parameters_fitted_by_Kasteren}
\begin{tabular}[tbp]{p{0.5cm}p{2.8cm}p{2.5cm}p{1cm}p{1.7cm}p{1cm}p{1cm}p{2.5cm}}
\\
\hline
Type & Activity & \multicolumn{2}{c}{Start time [min. from 0:00]} & Daily freq. & \multicolumn{2}{c}{Duration [min.]} & occurrence prob.\\
\cline{3-4}\cline{6-7}
& & $\mu_{a, s}$ & $\sigma_{a, s}$ & $\lambda_{a}$ & $\mu_{a, d}$ & $\sigma_{a, d}$ & $p_{a}$\\ \hline
\textbf{F} & \activity{prepare breakfast} & ~~562.74 (09:23) & ~~38.43 & - & ~~~~3.34 & ~~~~2.55 & -\\
           & \activity{take shower} & ~~619.02 (10:19) & 123.10 & - & ~~~~9.65 & ~~~~2.64 & -\\
           & \activity{prepare dinner} & 1159.78 (19:20) & ~~51.46 & - & ~~33.31 & ~~19.54 & -\\
           & \activity{go to bed} & 1416.14 (23:36) & ~~73.63 & - & 556.07 & ~~77.86 & -\\ \hline
\textbf{N} & \activity{use toilet} & - & - & 4.33 & ~~~~1.80 & ~~~~1.74 & -\\\hline
\textbf{R} & \activity{get drink} & - & - & - & ~~~~0.95 & ~~~~1.32 & 0.06\\
           & \activity{leave house} & - & - & - & 436.60 & 213.00 & 0.13\\
           & \activity{other} & - & - & - & ~~22.20 & ~~35.23 & 0.82\\
\hline
\end{tabular}
\end{table}

\begin{table}
\centering
\caption{Parameters of CASAS Aruba dataset in three types of activities (F: Fundamental, N: Necessary, R: Random). Here, $\mu_{a, s}$ and $\sigma_{a, s}$ denote the mean and the standard deviation of start time, $\mu_{a, d}$ and $\sigma_{a, d}$ denote those of duration time, and $\lambda_{a}$ is the mean frequency in a day, $p_{a}$ is an occurrence probability among random activities. The unit is a minute.}
\label{tab:parameters_fitted_by_Aruba}
\begin{tabular}[tbp]{p{0.5cm}p{2.8cm}p{2.5cm}p{1cm}p{1.7cm}p{1cm}p{1cm}p{2.5cm}}
\\
\hline
Type & Activity & \multicolumn{2}{c}{Start time [min. from 0:00]} & Daily freq. & \multicolumn{2}{c}{Duration [min.]} & occurrence prob.\\
\cline{3-4}\cline{6-7}
& & $\mu_{a, s}$ & $\sigma_{a, s}$ & $\lambda_{a}$ & $\mu_{a, d}$ & $\sigma_{a, d}$ & $p_{a}$\\ \hline
\textbf{F} & \activity{take breakfast} & ~~582.11 (09:42) & 58.27 & - & ~~10.04 & ~~~~9.56 & -\\
           & \activity{take lunch} & ~~805.98 (13:25) & 86.38 & - & ~~14.47 & ~~12.00 & -\\
           & \activity{take dinner} & 1119.39 (18:39) & 92.23 & - & ~~14.40 & ~~11.71 & -\\
           & \activity{sleeping} & ~~~~~~4.83 (00:04) & 66.65 & - & 427.16 & 110.77 & -\\
\hline
\textbf{N} & \activity{housekeeping} & - & - & 0.23 & ~~24.67 & ~~22.57 & -\\
           & \activity{meal preparation} & - & - & 4.86 & ~~13.40 & ~~11.96 & -\\
           & \activity{nap} & - & - & 0.14 & ~~85.75 & ~~68.87 & -\\
           & \activity{wash dishes} & - & - & 0.39 & ~~~~8.52 & ~~~~8.50 & -\\
\hline
\textbf{R} & \activity{go out} & - & - & - & 144.13 & ~~93.04 & 0.08\\
           & \activity{other} & - & - & - & ~~32.55 & ~~35.88 & 0.42\\
           & \activity{relax} & - & - & - & ~~66.15 & ~~74.55 & 0.46\\
           & \activity{work} & - & - & - & ~~26.53 & ~~29.31 & 0.04\\
\hline
\end{tabular}
\end{table}

\end{document}